\documentclass[10pt,journal,compsoc]{IEEEtran}
\usepackage{ifpdf}
\usepackage{hyperref}
\usepackage{cite}
\usepackage{graphicx}  
\usepackage{float}  
\usepackage{subfigure}  
\usepackage{amsthm}
\usepackage{amsmath}
\usepackage{amsfonts,amssymb}
\usepackage{booktabs}
\usepackage{CJK}
\usepackage{multirow}
\usepackage{tabularx}
\usepackage{tikz,hyperref}

\usepackage[ruled,linesnumbered]{algorithm2e}

\hypersetup{hidelinks}

\newtheorem{theorem}{Th.}[section]  

\ifCLASSOPTIONcompsoc
\else
  \usepackage{cite}
\fi
\ifCLASSINFOpdf
\else
\fi
\begin{document}
%
\title{Selecting task with optimal transport self-supervised learning for few-shot classification}
%
%
%
%

\definecolor{lime}{HTML}{A6CE39}
\DeclareRobustCommand{\orcidicon}{%
    \begin{tikzpicture}
    \draw[lime, fill=lime] (0,0)
    circle [radius=0.16] 
    node[white] {{\fontfamily{qag}\selectfont \tiny ID}}; \draw[white, fill=white] (-0.0625,0.095)
    circle [radius=0.007]; \end{tikzpicture}
    \hspace{-2mm}}
\foreach \x in {A, ..., Z}{%
    \expandafter\xdef\csname orcid\x\endcsname{\noexpand\href{https://orcid.org/\csname orcidauthor\x\endcsname}{\noexpand\orcidicon}}
    }

\newcommand{\orcidauthorA}{0000-0002-5388-9080}
\newcommand{\orcidauthorB}{0000-0002-1408-5514}
\newcommand{\orcidauthorC}{0000-0002-6691-6762}
\newcommand{\orcidauthorD}{0000-0001-6487-3183}

\author{Renjie~Xu,
        Xinghao~Yang\orcidD{},~\IEEEmembership{Member,~IEEE,}
        Baodi~Liu\orcidB{},~\IEEEmembership{Member,~IEEE,}\\
        Kai~Zhang\orcidC{},~\IEEEmembership{Member,~IEEE,}
        and~Weifeng~Liu\orcidA{},~\IEEEmembership{Senior Member,~IEEE}
\thanks{Renjie Xu is with the College of Oceanography and Space Informatics, China University of Petroleum (East China), Qingdao 266580, China(e-mail:xurj1998@163.com).}
\thanks{Weifeng Liu, Xinghao Yang and Baodi Liu are with the College of Control Science and Engineering, China University of Petroleum (East China), Qingdao 266580,
China (e-mail: liuwf@upc.edu.cn; yangxh\_upc@126.com; thu.liubaodi@gmail.com).}
\thanks{Kai Zhang is with the school of Petroleum Engineering, China University of Petroleum (East China), Qingdao 266580, China(e-mail: zhangkai@upc.edu.cn).}
\thanks{Corresponding Author: Weifeng Liu, Xinghao Yang}}

\IEEEtitleabstractindextext{%
\begin{abstract}
Few-Shot classification aims at solving problems that only a few samples are available in the training process. Due to the lack of samples, researchers generally employ a set of training tasks from other domains to assist the target task, where the distribution between assistant tasks and the target task is usually different.
To reduce the distribution gap, several lines of methods have been proposed, such as data augmentation and domain alignment. However, one common drawback of these algorithms is that they ignore the similarity task selection before training. The fundamental problem is to push the auxiliary tasks close to the target task.
In this paper, we propose a novel task selecting algorithm, named Optimal Transport Task Selecting (OTTS), to construct a training set by selecting similar tasks for Few-Shot learning. Specifically, the OTTS measures the task similarity by calculating the optimal transport distance and completes the model training via a self-supervised strategy. By utilizing the selected tasks with OTTS, the training process of Few-Shot learning become more stable and effective. Other proposed methods including data augmentation and domain alignment can be used in the meantime with OTTS. We conduct extensive experiments on a variety of datasets, including MiniImageNet, CIFAR, CUB, Cars, and Places, to evaluate the effectiveness of OTTS. Experimental results validate that our OTTS outperforms the typical baselines, i.e., MAML, matchingnet, protonet, by a large margin (averagely 1.72\% accuracy improvement).
\end{abstract}

}

\maketitle

\IEEEdisplaynontitleabstractindextext

%
\IEEEpeerreviewmaketitle

\IEEEraisesectionheading{\section{Introduction}\label{sec:introduction}}

%
%
%
%

\begin{figure}[t]
    \centering
    \includegraphics[scale=0.5]{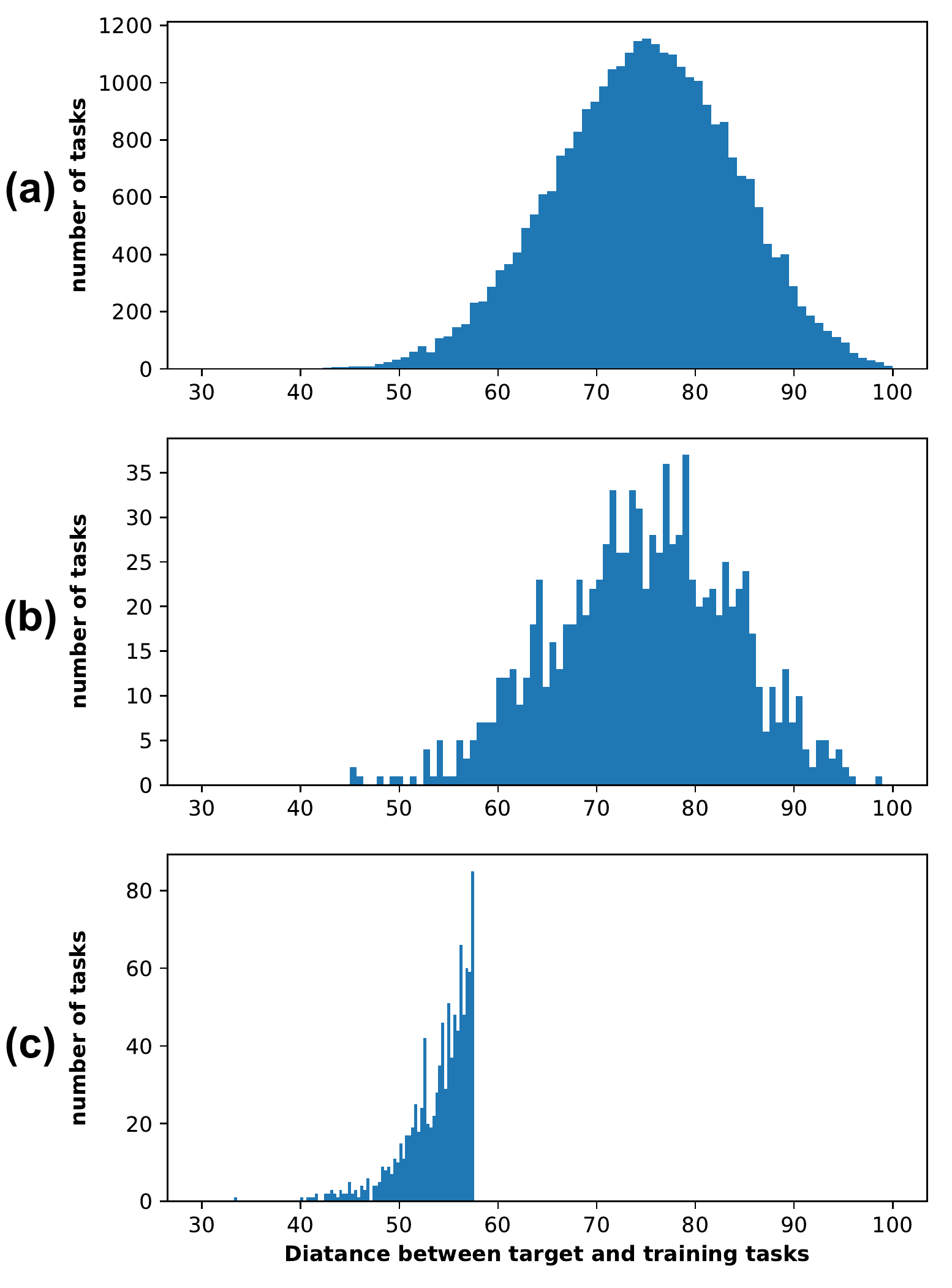}
    \caption{\textbf{(a)} shows the optimal transport distance (or similarity) between 30000 training tasks with another 1 target task. All of the tasks are from the MiniImageNet dataset. It can be seen that the distribution of the distance is approximate to a Gaussian distribution. \textbf{(b)} is the distribution of 1000 tasks that are randomly selected from the 30000 tasks. The most randomly selected tasks have low similarity with the target task. \textbf{(c)} is the distance distribution between the target task and the tasks selected by our OTTS. The shorter distance means higher similarity, which can benefit the training of the classifier.}
    \label{difference}
\end{figure}

\IEEEPARstart{F}{ew-Shot} learning \cite{few_shot_survey,protonet,TKDE_few_shot} targets to solve the problem that only a few samples can be accessed in the training process. Because of this property, the recent Few-Shot learning algorithms utilize samples from other tasks to assist training. In the real world, determining which tasks share the similar distribution with the target task is hard work. Therefore, how to select the assist tasks similar to the target task is a common problem. Training models with unsimilar tasks can bring the difficulty of domain adaption and cause lower accuracy. In this case, selecting tasks (shown in Fig. \ref{difference}) for training is a significant strategy for the target Few-Shot tasks than the existing methods, including data augmentation and domain alignment.

Data augmentation \cite{Data_Augmentation,Data_Augmentation_survey} generates new samples to increase diversity and quantity of training samples by using the existing samples. With the increasing of diversity of the training samples, the distribution between the new samples and the target samples might be similar. In another word, by using data augmentation, generalization ability of the model could be increased, because the data augmentation may generate some new samples that are similar to the samples in target tasks. Therefore, the augmentation effect is not steady and using similar samples can reduce the risk of invalid of data augmentation. Even if the data augmentation can minimize the difference between target tasks and generate new samples, the lack of similar training samples is still an ineluctable question. 

Domain alignment \cite{domain_generalization_survey,GOT_DA} empowers the model ability by leveraging training tasks from other domains. With domain alignment, the model learns the distribution of samples for both training and target tasks. Then it trains the feature extractor to make the distributions similar. The model can train a classifier with the source domains which have adequate training samples. However, the domain alignment also has some limits. Firstly, the properties of domain alignment tightly connect with the distribution of samples. For a Few-Shot task, the distributions of task samples are impossible to obtain. Secondly, domain alignment also requires the similarity between the training and target tasks. To achieve this, researchers manually select similar domains as the source and target domains. Intuitively, it is hard work for humans to select a training domain for each target task.

Therefore, it is an urgent demand to automatically select training tasks, which are similar to the target task. We can solve the Few-Shot problems \cite{few_shot_survey,protonet} which has no certain training tasks if we can select the training tasks automatically for target tasks. The similarity between tasks is the key point of selecting tasks. Researchers have proposed some methods to measure the distance between tasks including measuring method with probe network \cite{task2vec, task2vec_2} and measuring method with distribution \cite{task_distribution_OT}. The probe network is the most common strategy in recent years. For example, Task2Vec \cite{task2vec} firstly utilizes a probe network which is trained on the task requiring measurement. Then it applies the embedding of the probe network parameters as the feature of the task and measures the distance of tasks with the feature. The measuring method with distribution \cite{task_distribution_OT} utilizes Gaussian distribution to fit the distribution of the task samples. Then it adopts the optimal transpose \cite{ot,ot_graph,GOT_DA} to compute the distance between distributions as the distance between tasks. Although the above mentioned two strategies can measure the distance with tasks, both kinds of methods have high requirements for the number of samples. The tasks with adequate samples have a lower necessity of selecting tasks to assist training. Especially, Few-Shot tasks, which urgently need assist training tasks, are unable to process with the existing methods.

In this case, we noticed Automatic Curriculum Learning (ACL) \cite{CL_1,CL_2,CL_3,CL_survey}, which is a learning method that can select samples with the difficulty of samples. The ACL can be roughly divided into three categories, including Self-paced learning \cite{self_paced,self_paced_2}, Transfer Teacher \cite{transfer_teacher_1, transfer_teacher_2}, and Reinforcement learning teacher \cite{Reinforce_teacher_1,Reinforce_teacher_2}. With the Self-paced learning methods, model can automatically selecting samples with difficulty for further training. While in Transfer Teacher methods, a teacher model which is trained in other domains is used to design the curriculum. The Reinforcement learning teacher employs reinforcement learning to measure the difficulty of each sample and sort them as curriculum. The existing Transfer Teacher methods show the probability of using a teacher model to sorts the samples with the difficulty. Similar with the transfer teacher models, we can select tasks for Few-Shot learning if we can train a teacher model to measure the similarity between task.

In this paper, we propose a novel task selecting method to solve the Few-Shot task-selecting problem based on optimal transport, which is called Optimal Transport Task Selecting (OTTS). As a Few-Shot problem, the relationships between tasks are more likely contained in the samples instead of the unknowable distribution. And the relationship between samples could be another major difference except for the distribution of samples. Therefore, OTTS measures the similarity of tasks with the relationship between the features of the samples. In the process of OTTS, we build a graph for each task, where the features of the sample are represented as the nodes and the distance between of features are denoted as the edge. After building the graph, a distance based on optimal transport is designed to measure the similarity between tasks. The training of the feature encoder is based on self-supervised comparative learning with the optimal transport distance, which forces the similar tasks to gather while the unsimilar tasks scatter. Then, the OTTS can select the tasks before or during the training of classifier, which is similar to the teacher model of curriculum learning \cite{CL_1,CL_2,CL_3,CL_survey}. In this paper, we conduct a property experiment to verify the correctness of the training process and the distance of tasks. We also conduct an application experiment for the usage of OTTS on Cross-Domain Few-Shot learning. In the properties experiment, OTTS shows its superiority in selecting similar tasks not only from the same dataset but also from other datasets. By utilizing the tasks selected by OTTS, the training process become smoother with higher accuracy and lower loss. The result of the application experiment validates that the OTTS can benefit the Cross-Domain Few-Shot learning by selecting appropriate training tasks. The accuracy of models including protonet and matchingnet on cub, cars and places data sets has been improved by 1.72\%.

The main contributions of this paper can be summarized as 4 points below:

\begin{itemize}
    \item We propose a novel training strategy based on self-supervised learning algorithm called OTTS that can detect the similarity between Few-Shot tasks. 
    
    \item We propose an application of graph optimal transport distance to measure the similarity of tasks. In the OTTS, this distance is used as loss function during training and selecting criterion. 
    
    \item We conduct experiments to exam the correctness and verify the property of the OTTS algorithm. Find out the similarity of tasks included in the datasets including CUB, Cars, Places, CIFAR, and MiniImageNet in the view of the network.
    
    \item We propose a usage of OTTS for the Cross-Domain Few-Shot learning method, and verify the effect of the usage. 
\end{itemize}

The subsequent paper is organized as follows. The related works or areas which are tightly connected are discussed in section 2. In section 3, the algorithm details and the theories are narrated and analyzed. We make a summary of the experiments in section 4, which totally includes three parts of a different experiment. Finally, a discussion is engaged in section 5 to conclude this paper and look forward to the future.


 

\begin{figure}[t]
    \centering
    \includegraphics[scale=0.73]{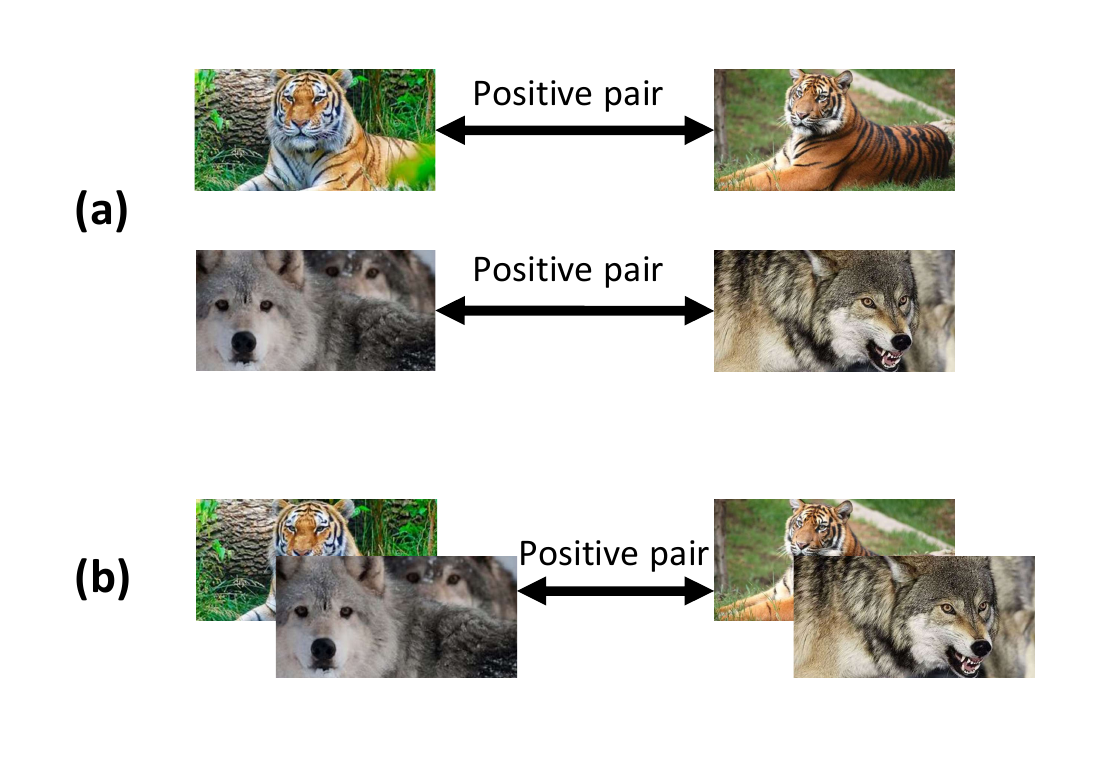}
    \caption{(a) is the positive pairs of traditional contrastive learning, which are samples from the same classes. (b) is the positive pairs in OTTS, which is the task containing the same classes samples.}
    \label{CSL}
\end{figure}

\begin{figure*}[t]
    \centering
    \includegraphics[scale=0.7]{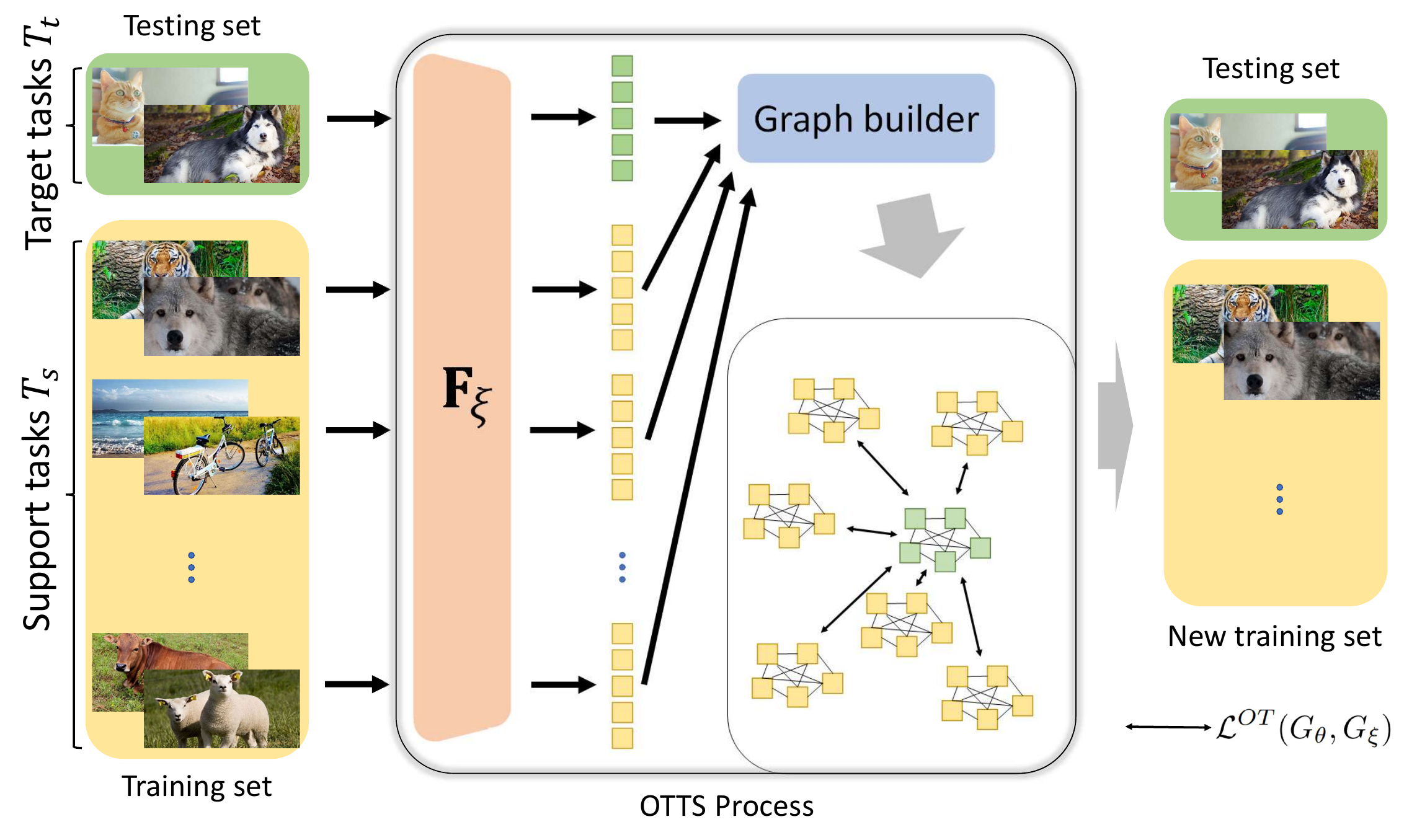}
    \caption{\textbf{Process of OTTS:} In OTTS, the samples of the tasks are input of the feature extractor $\mathbf{F}_\xi$. The extracted features are built into graphs with the Graph Builder with the relationship of the features. The distance between each graph of target tasks and each graph of the source model is computed by the optimal transport distance. And OTTS would select the most similar tasks with target task for the further training of the classifier.}
    \label{OTTS_using}
\end{figure*}

\section{Related works}

In this section, we introduce the existed works or fields which are most related to OTTS. The section contains four parts, i.e., optimal transport distance, self-supervised learning, Transfer Teacher model and Cross-Domain Few-Shot learning.

\subsection{Optimal Transport Distance}

Optimal transport \cite{ot,ot_graph,GOT_DA} is a method that aims at solving problems of the “moving simultaneously several items (or a continuous distribution thereof) from one configuration onto another” with the least possible effort \cite{ot}. In 1998, Rubner \emph{et al.} \cite{WD} proposed a novel distance which can be used in measuring the similarity of the images’ color distribution called Wasserstein distance based on optimal transport. 
\begin{equation}
    d_W = \min_{T\in\mathbf{T}}\frac{\sum_{i=1}^n\sum_{j=1}^m c_{ij}T_{ij}}{\sum_{i=1}^n\sum_{j=1}^m T_{ij}}
\end{equation}
In this equation, $T \in \mathbb{R}^{n\times m}$ is the flow that represents the number of supplies from the “supplier” $\alpha$ to the “consumers” $\beta$, $c$ is the cost, and $T_{ij}$ is the transport between $\alpha_i$ and $\beta_j$. Wasserstein distance can measure the distance between two distributions, but can only compute the distance between samples to samples. To measure the distance between relationship to relationship, Gromov-Wasserstein distance \cite{GWD} is proposed:
\begin{align}
    d_{GW} &= \inf_{T\in\mathbf{T}} \mathbb{E}_{T_{ij} \in T,T_{i'j'} \in T}[\mathcal{L}(\alpha,\beta,\alpha',\beta')] \notag\\
    &=\min_{T\in\mathbf{T}} \sum_{i,i',j,j'} T_{ij}T_{i'j'}\mathcal{L}(\alpha_i,\beta_j,\alpha'_{i'},\beta'_{j'})
\end{align}

In Eq. (2), the $\mathbf{T}$ is the set of flows which contains all the possible flows from $T:\alpha \to \beta$. $\mathcal{L}(\alpha_i,\beta_j,\alpha'_{i'},\beta'_{j'})=||c(\alpha_i,\alpha'_{i'})-c(\beta_j,\beta'_{j'})||$ is the norm of the difference between the cost $c(\alpha_i,\alpha'_{i'})$ and $c(\beta_j,\beta'_{j'})$. By using this distance, the similarity between the relationship of $\alpha_i$, $\alpha'_{i'}$ and the relationship of $\beta_j$, $\beta'_{j'}$ could be computed, which are the relationship between the nodes of two different graphs.

In recent years, optimal transport distance has been deployed in many areas. Muzellec \emph{et al.} \cite{ot_distribution_1} proposed a probabilistic embedding method which uses optimal transport distance to compare the distributions. Frogner \emph{et al.} \cite{ot_distribution_2} designed similar OT distance to measure the discrete measures. In 2019, Graph Optimal Transport (GOT) \cite{ot_graph} employed WD and WGD to measure the difference between two graphs. Furthermore, Chen \emph{et al.} \cite{GOT_DA} utilized the GOT to alignment two graphs which belong to different domains. Following GOT \cite{ot_graph,GOT_DA}, we utilize GOT to measure the distance between two tasks.

\subsection{Contrastive Self-Supervised Learning}

Contrastive self-supervised learning \cite{SS_survey}, which belongs to unsupervised learning, completes the training process by utilizing the training samples information rather than the label information. This method is especially valuable in the situation that plenty of unlabeled samples are available in the training phase. Contrastive methods \cite{moco,BYOL} are commonly used to assemble the same image (“positive pairs”) together and simultaneously scatter the different ones (“negative pairs”).

A slow-moving average target network \cite{BYOL} designs a mixing process to update the parameter of the target network, which is used in many self-supervised learning models. This mixing process can produce stable output of the target network. The mixing process is commonly represented as
\begin{align}
    \xi &\gets \tau\xi + (1-\tau)\theta, \tau\in(0,1)
\end{align}
where $\xi$ is the parameter of the target network which is slow-moving. The $\theta$ is a set of parameter to update the target network.

Bootstrap Your Own Latent (BYOL) \cite{BYOL} is a typical method of contrastive learning. In BYOL, the feature $Z_\theta$ is extracted from the training sample $X$ by the online model $\mathbf{F}_\theta$ with parameter $\theta$. The feature $Z_\theta$ is used to predict the feature $Z_\xi$ extracted by the target model $\mathbf{F}_\xi$. After obtaining the similarity between $Z_\theta$ and $Z_\xi$ as the contrastive loss, an optimizer is used to update the parameters of the online model and the parameter of the target model is updated with a mixing process.
\begin{align}
    \theta &\gets optimizer(\theta, \nabla_\theta \mathcal{L}^{BYOL}_{\theta,\xi}, \eta),\\
    \xi &\gets \tau\xi + (1-\tau)\theta
\end{align}
where $\mathcal{L}^{BYOL}_{\theta,\xi}$ is the contrastive loss used in BYOL, $\eta$ is the target decay rate.

In existed self-supervised learning methods, the samples normally used in the training process are the unlabeled images, sound signals, or sentences. Therefore a natural thought is that the tasks can be processed in a similar way if we can extract their effective feature. Following the same process as contrastive self-supervised learning, we can determine which tasks are similar like determining the similarity of samples as shown in Fig. \ref{CSL}. 

\subsection{Transfer Teacher Model}

Transfer Teacher is a popular method in curriculum learning. In Transfer Teacher methods, a strong teacher model is pretrained to measure and sort the difficulty of samples with the performance of the teacher. In existed works, the teacher model could be a larger model than the student \cite{transfer_teacher_1} or have the same structure with student model \cite{transfer_teacher_2}. Generally, the teacher models are trained with a large enough dataset to obtain knowledge of label predicting and thus compute the loss of each sample. It can be recognized that the difficulty of the sample is based on the prediction of teacher and label.

With the optimal transport distance and self-supervised learning method mentioned above, we propose OTTS, a task selecting method trained with the self-supervised method, and optimal transport distance by regarding few-shot tasks as graphs. The main idea of the training process of OTTS is similar to self-supervised learning but on the plane of tasks. In OTTS, the positive pairs are tasks which include the same classes of samples. By training with the self-supervised method, OTTS can assemble similar graphs which represent the “features” of similar tasks, and scatter the unsimilar ones. The major difference lies that the teacher models measure the difficulty of samples, while OTTS measures the similarity between tasks.

\subsection{Cross-Domain Few-Shot Learning}

Cross-Domain Few-Shot (CDFS) Learning \cite{CDFS_1,CDFS_2} is a novel field with increasing popularity recent years. In most CDFS learning methods, researchers adopt the domain generation \cite{domain_generalization_survey} or transfer learning \cite{transfer} methods to force the classifier model to adapt to a new domain. In the experimental part of this paper, we use training tasks selected by OTTS to decrease the difficulty of adaptation and improve the performance of existing models.

\section{Optimal transport task selection}

In this section, we introduce the details of OTTS method. We introduce the definition of the task selecting problem and the structure of OTTS in the subsection 3.1 and 3.2, respectively. Then we describe the training strategy of the feature extractor in the subsection 3.3. In subsection 3.4, we introduce the optimal transport loss function, and make a short prove for the relationship between the optimal transport distance and the loss of classifier.

\subsection{Problem Definition}

In a Few-Shot task selecting problem, the model need to select a set of support tasks $\{\mathrm{T}^{n_i}_{S}\}_{i}$ from tasks $\{\mathrm{T}^i_{S}\}_i$ in source domain $\mathrm{D}_S$ according to a testing set of target tasks $\{\mathrm{T}_T\}$, where ${n_i}$ is the selected task "ID" in the $\mathrm{T}_S$. Particularly, each support task is represented as $\mathrm{T}_S=\{(x_s^1,y_s^1),...,(x_s^N,y_s^{N*K}),(x_{s}^{N*K+1},\cdot),...,(x_{s}^{N*K+M},\cdot)\}$ and target task is represented as $\mathrm{T}_T=\{(x_t^1,y_t^1),...,(x_t^N,y_t^{N*K}),(x_{t}^{N*K+1},\cdot),...,(x_{t}^{N*K+M},\cdot)\}$. In this case, each task contains K labeled samples for each of the N classes, which is commonly called $N$-way $K$-shot task. We assume the target tasks are from a unknowable target domain $\mathrm{D}_T$. The task-selecting model targets to select tasks from the source domain $\mathrm{D}_S$ which are similar to the target task. By doing this, the OTTS model selects a set of training tasks in which samples are most possibly have the same features or distribution with the target tasks shown in \ref{OTTS_using}.

\subsection{Structure of OTTS}

The structure of OTTS includes a feature extractor $\mathbf{F}_\xi$ and a graph builder. The feature extractor is designed to extract the feature of each sample of a task. Then the graph builder can graph with the features. Each graph in this process represents the feature of a task, where the nodes donate the samples feature $Z$ and the edges represent the relationship between samples. The feature extractor is the trainable part of OTTS while the output of the graph builder only depends on the similarity between the features. In the training process, we also use a online model $\mathbf{F}_\theta$ which has the same structure with the feature extractor to assist training. 

\subsection{Training Strategy of the Feature Extractor}

\begin{figure*}[t]
    \centering
    \includegraphics[scale=0.7]{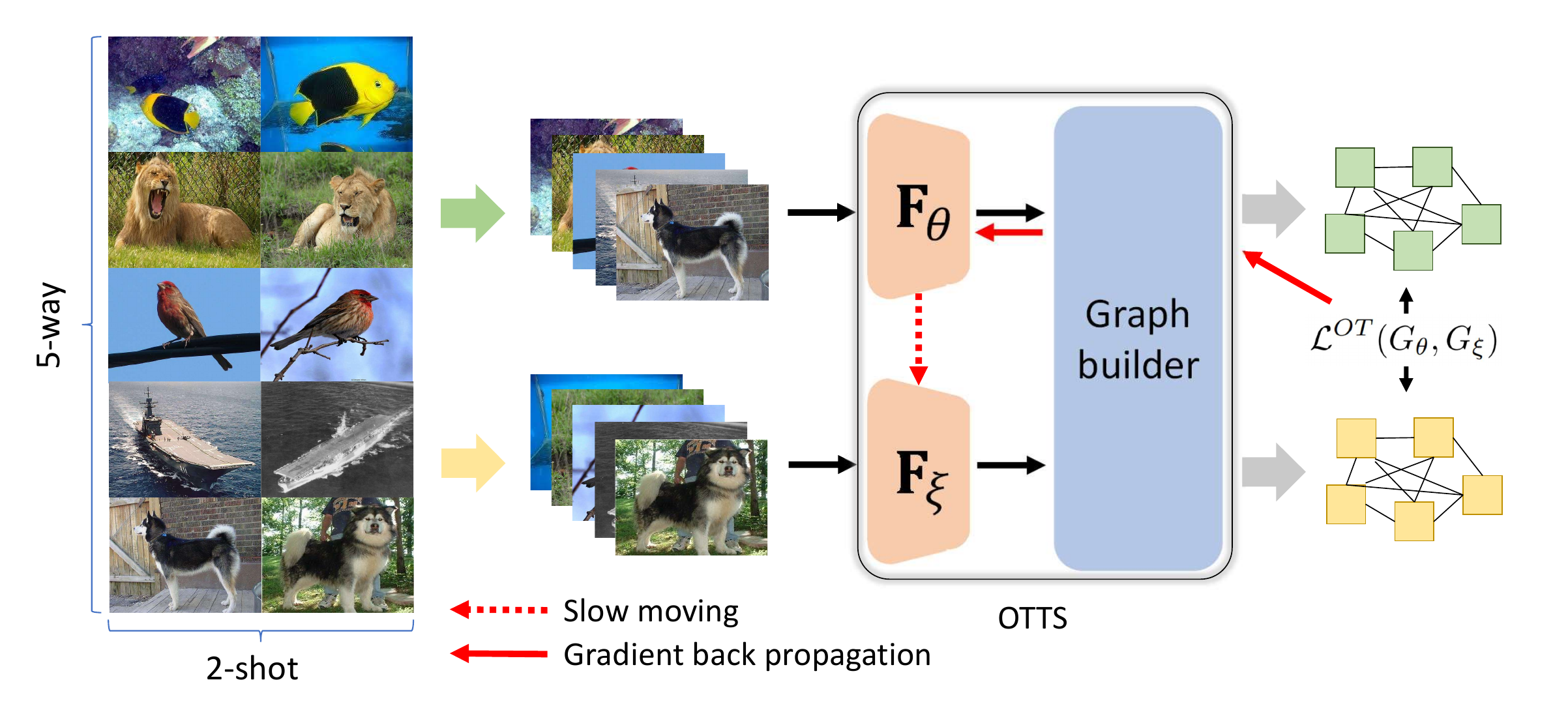}
    \caption{\textbf{Training OTTS:} The self-supervised training method is used in the training process. We saw the graphs which construct by the samples from the same task as the multi-view of the same task. We use the self-supervised training method to align the different views of the same task. To do so, the OTTS can obtain a set of parameters for feature extractor $\mathbf{F}_\xi$ to extract the feature of the tasks.}
    \label{OTTS_training}
\end{figure*}

During the training process, we use a self-supervised training strategy like BYOL but on the planes of tasks. As a task selecting problem, what we need during the training process is a huge number of tasks but not samples. To train OTTS, we prepare $N$-way $2$-shots tasks and split each of the tasks into two $N$-way $1$-shot tasks $T'$ and $T''$. The samples of $N$-way $1$-shot tasks are fed as the input for the online and target models $\mathbf{F}_\theta$, $\mathbf{F}_\xi$ separately as a pair of positive samples. 

The features $\{Z\}_{i=1}^N$ which extracted from each of these samples are then constructed into graphs with the graph builder. In this step, the relationship between samples of the tasks is computed and can be used in the calculation of the loss function. For each N-way 1-shot task, we build a graph $G$ with N nodes to represent the “feature of the task”. We consider each pair of the features of the tasks as a pair of different representations of the same tasks. Intuitively, the OTTS model can find the same information between the representation and ignore the differences. 

Therefore, in the training process of OTTS, we build an online model $\mathbf{F}_\theta$ which is the same as the feature extractor to assist training and the feature extractor is seen as the target model $\mathbf{F}_\xi$. We employ the graph $G_\theta$ of task feature ${Z_\theta}$ which is extracted by the online model $\mathbf{F}_\theta$ to predict the graph $G_\xi$ of feature ${Z_\xi}$ extracted by the target model $\mathbf{F}_\xi$. Then exchange the order of the tasks to obtain another pair of graphs $G'_\theta$ and $G'_\xi$. 
\begin{align}
    \mathcal{L}^{OTTS} = \mathcal{L}^{OT}(G_\theta,G_\xi)+\mathcal{L}^{OT}(G'_\theta,G'_\xi)
\end{align}
And updating the model with the with optimal transport loss function below
\begin{align}
    \theta &\gets optimizer(\theta, \nabla_\theta \mathcal{L}^{OTTS}, \eta),\\
    \xi &\gets \tau\xi + (1-\tau)\theta
\end{align}
where the $\mathcal{L}^{OT}(G_\theta,G_\xi)$ is the optimal transport loss function. The training strategy is shown in Fig. \ref{OTTS_training}.

\subsection{Optimal Transport Loss Function}

The optimal transport loss function used in this paper is the summation of Wasserstein distance and the Gromov-Wasserstein distance. As mentioned above, each task is embedded as a graph, which can be viewed as discrete measures in the graph space. Therefore, the distance between two graphs can be measured by Wasserstein distance:
\begin{align}
    \mathcal{L}_{W}(G_\theta,G_\xi) = &\min_{T(G_\theta,G_\xi)\in\mathbf{T}}\sum_{i=1}^N\sum_{j=1}^M c_{ij}T_{ij}\notag\\
    = &\min_{T(G_\theta,G_\xi)\in\mathbf{T}} \sum_{T_{ij}\sim T}c_{ij}\notag\\
    = &\min_{T(G_\theta,G_\xi)\in\mathbf{T}} \sum_{T_{ij}\sim T} ||Z_\theta^i - Z_\xi^j||
\end{align}
where $c_{ij} = ||Z_\theta^i - Z_\xi^j||$ is the cost of transport from the feature $Z_\theta^i$ in graph $G_\theta$ and $Z_\xi^i$ in graph $G_\xi$. The Wasserstein distance can compute the similarity between the node of two graphs, which represent the feature of samples. During the training process of the classifier, the classifier loss of each task should be the same if the tasks are similar. The reason is that similar tasks theoretically share the same classifier parameters. Thus we can prove the relationship between the optimal transport distance and the similarity of tasks for classifier:

\begin{theorem}
Given two graphs $G_\theta, G_\xi$ of sample feature from different tasks. ${Z_\theta} \in G_\theta$ and ${Z_\xi} \in G_\xi$ are the features in the graph. For $\forall \epsilon \in \mathbb{R}^+, \exists \delta$ such that for all $G_\theta, G_\xi$, 
\begin{align}
    \mathcal{L}_{W}(G_\theta,G_\xi)= \min_{T(G_\theta,G_\xi)\in\mathbf{T}}\sum_{i=1}^N\sum_{j=1}^M c_{ij}T_{ij}<\delta
\end{align}
implies that
\begin{align}
    |p(y=Y|G_\theta,w) - p(y=Y|G_\xi,w)|<\epsilon
\end{align}
where $w$ is the parameter of the model.

\textit{Proof}: Given $c_{ij}$ is the cost that transport from $Z_\theta^i$ to $Z_\xi^j$. For $T(G_\theta,G_\xi)\in\mathbf{T}$ that minimize the loss.
\begin{align}
    &|p(y=Y|G_\theta,w) - p(y=Y|G_\xi,w)| \notag\\
    =&|\prod_i p(y^i|Z_\theta^i,w) - \prod_{T_{ij}\sim T(G_\theta,G_\xi)} p(y^i|Z_\xi^j,w)| \notag\\
    =&|\prod_i p(y^i|Z_\theta^i,w) - \prod_{T_{ij}\sim T(G_\theta,G_\xi)} p(y^i|Z_\theta^i,w)p(Z_\theta^i|Z_\xi^j)| \notag\\
    =&|1-\prod_{T_{ij}\sim T(G_\theta,G_\xi)} p(Z_\theta^i|Z_\xi^j)|\prod_i p(y^i|Z_\theta^i,w)
\end{align}
The distribution of $Z_\xi^j$ with noise is a Gaussian distribution \cite{GPML} with $\mu = Z_\xi^j$:
\begin{align}
    p_{Z_\xi^j}(x)= \mathcal{N}(Z_\xi^j,\Sigma^j)
\end{align}
The possibility for $x = Z_\theta^i$ is
\begin{align}
    &p(Z_\theta^i|Z_\xi^j)
    =p_{Z_\xi^j}(Z_\theta^i)\notag\\
    =&(2\pi)^{-\frac{D}{2}}|{\Sigma^j}|^{(-1/2)}exp(-\frac{1}{2}(Z_\theta^i-Z_\xi^j)^T{\Sigma^j}^{-1}(Z_\theta^i-Z_\xi^j))
\end{align}
Set $\lambda^j_{max}$ is the max value of the characteristic value of the positive definite matrix ${\Sigma^j}^{-1}$, for $c(Z_\theta^i,Z_\xi^i)=(Z_\theta^i-Z_\xi^j)^T(Z_\theta^i-Z_\xi^j)^{1/2}$
\begin{align}
     &(2\pi)^{-\frac{D}{2}}|{\Sigma^j}|^{(-1/2)}exp(-\frac{1}{2}(Z_\theta^i-Z_\xi^j)^T{\Sigma^j}^{-1}(Z_\theta^i-Z_\xi^j)) \notag \\
     > &(2\pi)^{-\frac{D}{2}}|{\Sigma^j}|^{(-1/2)}exp(-\frac{1}{2}{\lambda^j_{max}}^{D}(Z_\theta^i-Z_\xi^j)^T(Z_\theta^i-Z_\xi^j)) \notag\\
     = &(2\pi)^{-\frac{D}{2}}|{\Sigma^j}|^{(-1/2)}exp(-\frac{1}{2}{\lambda^j_{max}}^{D}c(Z_\theta^i,Z_\xi^i)^2)
\end{align}
Substitute (16) into (13), $\lambda_{max}=\max_j \lambda^j_{max}$. 
\begin{align}
    &|1-\prod_{T_{ij}} (2\pi)^{-\frac{D}{2}}|\Sigma|^{-\frac{1}{2}}e^{-\frac{1}{2}{\lambda^j_{max}}^{D}c(Z_\theta^i,Z_\xi^i)^2}|\prod_i p(y^i|Z_\theta^i,w)\notag\\
    <&|1- K e^{-\frac{1}{2}{\lambda_{max}}^{D}\sum_{T_{ij}}c(Z_\theta^i,Z_\xi^i)^2}|\prod_i p(y^i|Z_\theta^i,w)\notag\\
    <&|1- K e^{-\frac{1}{2}{\lambda_{max}}^{D}(\sum_{T_{ij}}c(Z_\theta^i,Z_\xi^i))^2}|\prod_i p(y^i|Z_\theta^i,w)\notag\\
    <&|1- K e^{-\frac{1}{2}{\lambda_{max}}^{D}(\delta)^2}|\prod_i p(y^i|Z_\theta^i,w)=\epsilon
\end{align}
where $K=((2\pi)^{-\frac{D}{2}}|\Sigma|^{-\frac{1}{2}})^j$. For every $\epsilon$, exist $\delta$ such that $\forall G_\theta, G_\xi$ $\sum_{T_{ij}\sim T}c_{ij}=\sum_{T_{ij}\sim T}c(Z_\theta^i,Z_\xi^i)<\delta$, the $|p(y=Y|G_\theta,w) - p(y=Y|G_\xi,w)|<\epsilon.$$\hfill\qedsymbol$
\end{theorem}

In this theory, we proved that if the distance between two tasks is small, the losses of the classifier with the same parameters for each task are also similar. The similar losses on each point would lead to the same solution of classifier parameters. Therefore, training with the tasks which cause similar losses can bring a stable and effective training process for the classifiers.

However, the Wasserstein distance can not express all the information contained in the graph. It can only measure the similarity between features. We hence employ the Gromov-Wasserstein distance to represent the relationship of nodes, which implies the information of the relationship between samples from different classes. Wasserstein distance can help the OTTS model confirm the position of each sample in the task, further correcting the similarity between tasks. The correcting effect has a positive relationship with the number of samples used in the graph.
\begin{align}
    &\mathcal{L}_{GW}(G_\theta,G_\xi)\notag\\
    = &\inf_{T\in\mathbf{T}} \mathbb{E}_{T_{ij} \in T,T_{ij} \in T}[\mathcal{L}(G_\theta^i,G_\theta^{i'},G_\xi^j,G_\xi^{j'})]\notag\\
    =&\min_{T\in\mathbf{T}} \sum_{i,i',j,j'} T_{ij}T_{i'j'}\mathcal{L}(Z_\theta^i,Z_\theta^{i'},Z_\xi^j,Z_\xi^{j'})
\end{align}
Therefore, the final loss is the sum of WD loss and GWD loss:
\begin{align}
    \mathcal{L}^{OT}(G_\theta,G_\xi) = r*\mathcal{L}_{W}+(1-r)\mathcal{L}_{GW}
\end{align}
where $r\in(0,1)$. With this optimal transport loss function, the $L_{GW}$ is more like a regularization term to further limit the structure of the graphs. By training with the loss function with the self-supervised learning method. The tasks which have similar features can be assembled. After the training process, the target model in self-supervised learning and a graph builder are able to extract the task features. The similarity between tasks is calculated with the same similarity function as the training loss. After computing the similarities between the target task and the source tasks, OTTS can select tasks from source tasks as needed.

The details of the training and application process of OTTS is shown in the Algorithm \ref{OTTS}.

\begin{algorithm}
\caption{The proposed OTTS algorithm}\label{OTTS}
\SetAlgoLined

\KwIn{A set of $N$-way 2-shot tasks $\{T\}$, Online model $\mathbf{F}_\theta$ and target model $\mathbf{F}_\xi$. A set of target tasks $\{\mathrm{T}_T\}$ and a large set $\{\mathrm{T}^i_S\}_i$ which includes a lots of tasks}

\KwOut{A set of training tasks $\{\mathrm{T}^{n_j}_S\}_j$}

/*\emph{The training process of OTTS}\qquad\qquad\qquad*/\\
\For{A batch of $T\in\{T\}$}{
    Divide the each $N$-way 2-shot tasks $T$ into 2 $N$-way 1-shot tasks $T'$, $T''$\;
    Extract the features $\{Z_\theta\}$ of samples in $T'$ by $\mathbf{F}_\theta$ and the feature $\{Z_\xi\}$ in $T''$ by $\mathbf{F}_\xi$\;
    Build graph $G_\theta$ with $\{Z_\theta\}$ and $G_\xi$ with $\{Z_\xi\}$\;
    Exchange $T'$ and $T''$ and build graphs $G'_\theta$ and $G'_\xi$ with the same process above\;
    Compute the optimal transport loss\\ $\mathcal{L}^{OTTS}= \mathcal{L}^{OT}(G_\theta,G_\xi) +  \mathcal{L}^{OT}(G'_\theta,G'_\xi)$\;
    Update the parameter of online model $\mathbf{F}_\theta$ with\\ $\theta \gets optimizer(\theta, \nabla_\theta \mathcal{L}^{OTTS}(G_\theta,G_\xi), \eta)$\;
    Update the parameter of target model $\mathbf{F}_\xi$ with\\ $\xi \gets \tau\xi + (1-\tau)\theta$\;
}
Gain a set of parameter $\xi$ for the target model\;
/*\emph{The application of OTTS}\qquad\qquad\qquad*/\\
Extract the features $\{Z^{n_i}_s\}_i$ of $\mathrm{T}_S\in\{\mathrm{T}^i_S\}_i$ with $\mathbf{F}_\xi$ and build a graph $G^i_s$ for each feature\;
Compute a set of optimal transport distance $\{\mathcal{L}^{OT}_i(G_t,G^i_s)\}_i$ between $G_t$ and $G^i_s\in\{G^i_s\}_i$\;
Sort the $\{\mathcal{L}^{OT}_i(G_t,G^i_s)\}_i$ in an increased order $\{\mathcal{L}^{OT}_{n_i}(G_t,G^{n_i}_s)\}_i$\;
Select the top M elements, i.e. $\{\mathcal{L}^{OT}_{n_i}(G_t,G^{n_i}_s)\}_{i=1}^M$ and the corresponding tasks $\{\mathrm{T}^{n_i}_S\}_{i=1}^M$\;
\Return  Top M elements of the sorted list $\{\mathrm{T}^{n_i}_S\}_{i=1}^M$.
\end{algorithm}

\section{Experiment}
In this section, we run experiments on several datasets which are commonly used in the Few-Shot and Cross-Domain method \cite{Cross_domain,Cross_domain_2,CDFS_1,CDFS_2} to verify the performance of OTTS on Few-Shot and Cross-Domain Few-Shot learning. 

\subsection{Dataset Description}
In this section, we introduced the datasets used in the experiments. We use datasets which are commonly used in few-shot learning and transfer learning to verify the effects of OTTS. The details of the datasets are shown below:

\emph{MiniImageNet:}
The MiniImageNet \cite{meta_learning_MiniImageNet} dataset is a subset of the ImageNet dataset. Specifically, the MiniImageNet contains totally 60000 images, belonging to 100 classes, which have 600 samples per class. The size of each image is 84*84. For fair comparison, we randomly divided the MiniImageNet into training, validation, and testing set with the proportion of 64:16:20.

\emph{CIFAR100:}
The CIFAR \cite{CIFAR} includes 100 classes of images. Each of the classes contains 600 samples of size 32*32. The samples in each class is divided into 500 training samples and 100 testing samples.

\emph{CUB:}
The CUB \cite{CUB} dataset is consist of 200 classes of birds images with totally 11788 samples. These samples are divided into a subset with 5994 samples for training and the rest 5794 samples for testing.

\emph{Cars:}
The Cars \cite{Cars} dataset contains 196 classes of images which are consist of 16185 images with different kinds of cars. The dataset is divided into a training set with 8144 images and testing classes with 8041 images. 

\emph{Places:}
The Places \cite{places} dataset comprises over 10 million images. The images are categorized into more than 400 unique scene classes. Each class includes 5000 to 30000 training images.

The size of the images in CIFAR100, CUB, Cars, and Places are resized to 84*84 to suit the size of images in MiniImageNet. 

\begin{figure*}[!t]
\centering
\subfigbottomskip=1pt 
\subfigcapskip=-5pt 
\subfigure[\textbf{CUB Vs Places}]{\includegraphics[width=165pt]{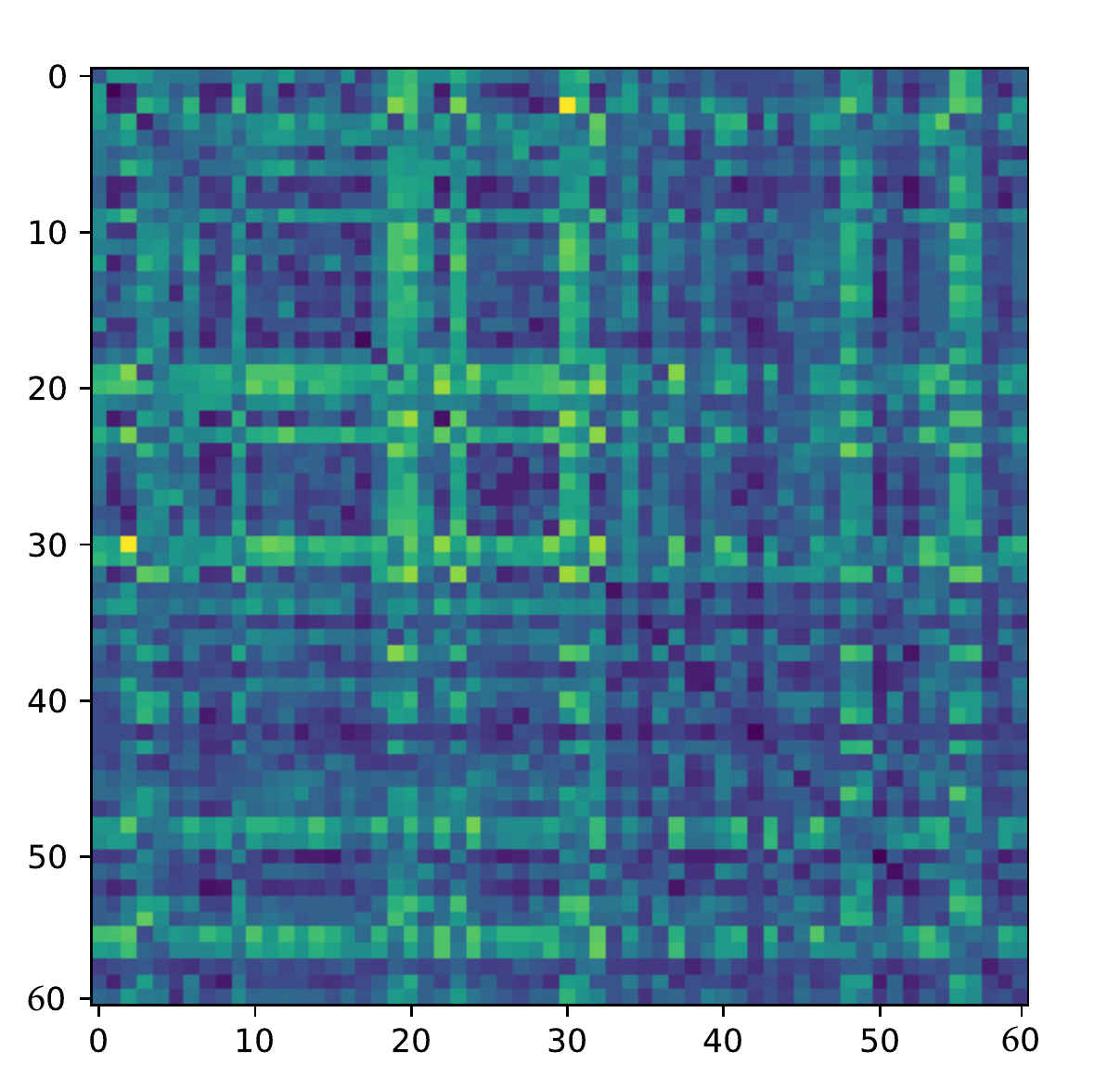}}
\subfigure[\textbf{Places Vs Cars}]{\includegraphics[width=165pt]{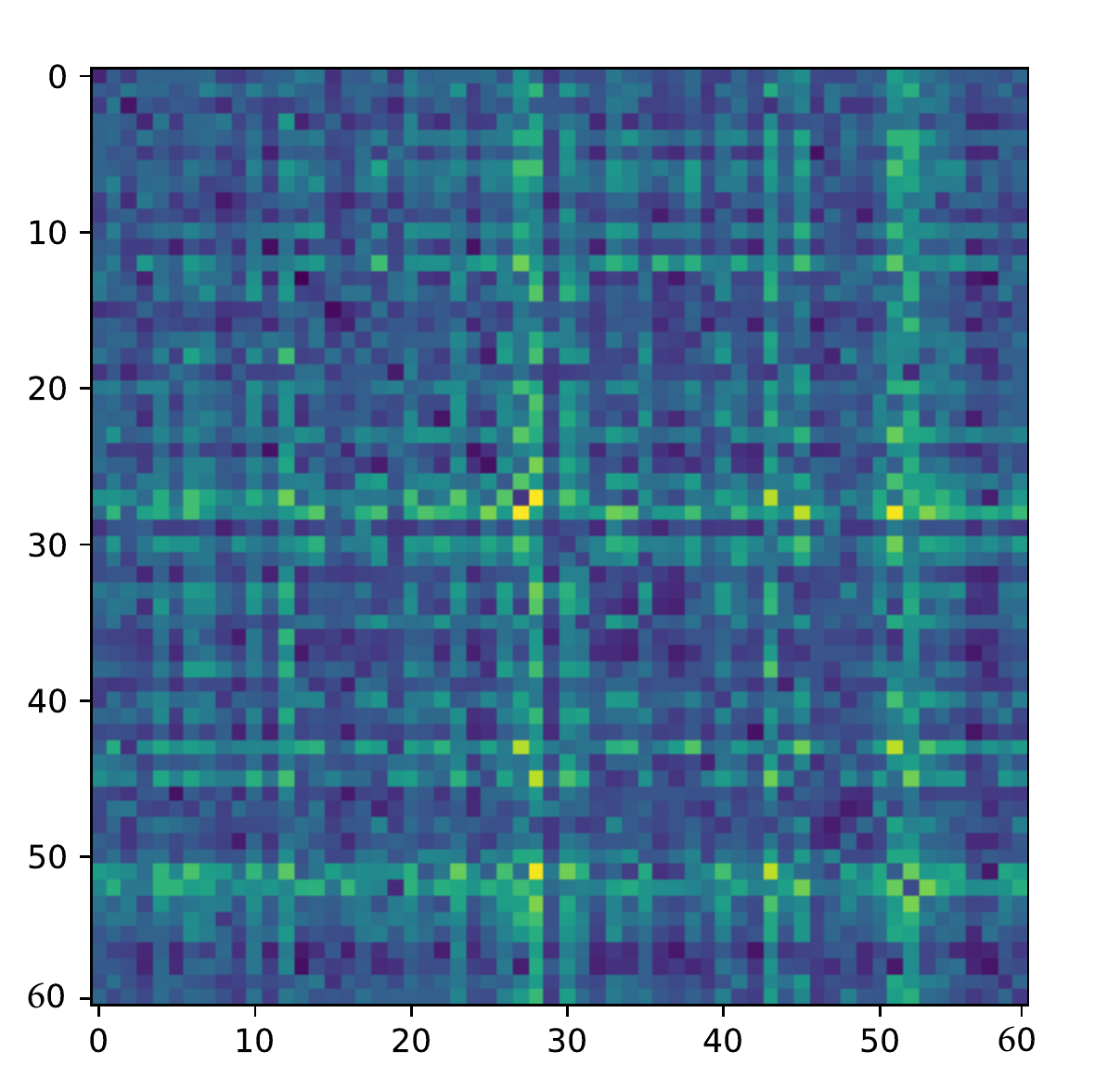}}
\subfigure[\textbf{Cars Vs CUB}]{\includegraphics[width=165pt]{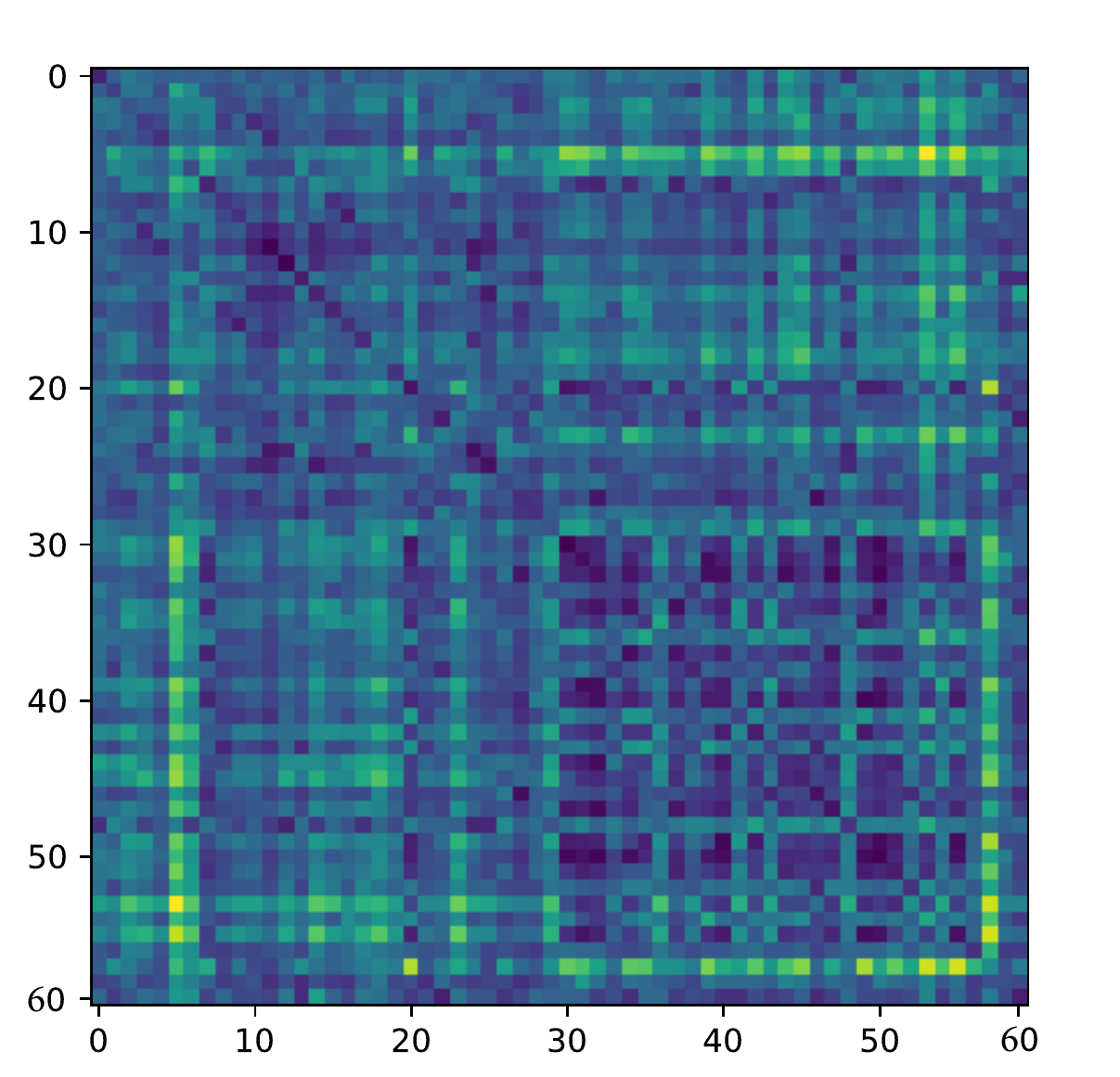}}
\caption{\textbf{Similarity between tasks cross domain}: This figure shows the similarity between different cross domains. The darker the color, the more similar the tasks are. Each domain contains 30 tasks, while the diagonal shows the similarity with itself.}
\label{distance_figure}
\end{figure*}

\subsection{Baseline Description}

We compare our method with typical baselines, including both metric-based Few-Shot models, i.e., matchingnet \cite{matchingnet}, protonet \cite{protonet} and optimization-based model, i.e., MAML \cite{MAML}.

\emph{Matchingnet:}
Matchingnet adopts pair-wise comparisons between the samples in the support set and the new input testing set. The predictions of the Matchingnet are based on the weighted similarity of samples in the support set and testing set. 

\emph{Protonet:}
Protonet predicts the label of the testing samples by calculating the distance between the feature of the testing sample and the prototype of each class. The prototype of each class is obtained by all the samples belonging to the class. 

\emph{MAML:}
MAML is a Model-Agnostic optimization-based model that uses inner and outer optimization procedure to obtain the second-order gradient. The key idea of MAML is to obtain the ability to fast optimize to new tasks. This ability is tightly related to the similarity between the training and testing tasks.

\subsection{Details of OTTS}

The training process of OTTS has been introduced above. Resnet-50 \cite{resnet} is used as the online and target model in the self-supervised training process. In the training process, we used 25600 5-way 2-shot tasks from the training set of MiniImageNet. In each epoch, we input a batch of tasks into the self-supervised model and compute the optimal transport loss of the batch. The size of each batch of tasks is 64. The optimizer used in this process is Adam \cite{adam}.

After training the OTTS model, we randomly select 30000 tasks from the training set of MiniImageNet as the training set. In the experiments below, these tasks are used if we need any tasks from the training set of MiniImageNet.

\subsection{Implementation Details}

In the experiments, we utilize both optimization-based and metric-based few-shot models include MAML \cite{MAML}, Matchingnet \cite{matchingnet}, and Protonet \cite{protonet} as the baselines. For MAML, we utilize a four-layer CNN (Conv4) as the backbone network. We set the learning rate $\alpha=0.02$ while the meta-learning rate $\beta=0.002$ of MAML. The model is trained for 10 epochs with 400 training tasks from the training set. After the training process of each epoch, we utilize 10 set of testing tasks from MiniImageNet or CIFAR. Each set includes 5 tasks from the testing set to evaluate the performance of the model. The other hyperparameters are the same as the setting in the paper of MAML. To compare the effect of OTTS, we utilize selected training tasks that are similar to testing tasks while the control group uses the random training tasks.

For the Matchingnet and Protonet, we also utilize a Conv4 as the backbone network. The training strategy of Matchingnet and Protonet is Feature-wise Transformation (FT) \cite{ex3}. Particularly, we train each model via 8 epochs while each epoch includes 10 couples of training tasks. The other settings are the same as the settings in FT \cite{ex3}. After training the model, we utilize a set of testing tasks from the testing set of the target domain (including CUB, Cars, and Places) to test the result of the training. We utilize the testing tasks and similar training tasks selected by OTTS and the random testing and training tasks to make comparisons. 

We adopt two quantitive measurements, i.e., classification accuracy and model loss, to evaluate the classification performance.

\subsection{Experimental Results}

In this section, we compare the OTTS enhanced baseline model with the basic baseline model to verify the property of the OTTS. Then we propose the application of the OTTS on the cross-domain few-shot classification learning.

\begin{table}[!t]
\renewcommand{\arraystretch}{1.8}
\caption{Distance between datasets.}
\label{distance_table}
\centering
\begin{tabular}{cccc}
\toprule
Dataset             &Average    &Max        &Min\\ 
\midrule
CUB                 &62.05      &133.57     &19.86\\
Cars                &64.32      &123.07     &18.06\\
Places              &66.01      &137.11     &19.74\\
\midrule
CUB vs Cars         &68.37      &137.17     &21.09\\
Cars vs Places      &65.23      &138.46     &23.79\\
Places vs CUB       &69.29      &141.12     &23.95\\
\midrule
CUB vs MiniImageNet &64.05      &136.13     &20.74\\
Cars vs MiniImageNet&68.01      &130.11     &25.18\\
Places vs MiniImageNet&67.70     &143.37     &24.36\\
\bottomrule
\end{tabular}
\end{table}

\subsubsection{Property experiment}

This experiment aims to verify the property of the OTTS method. In this experiment, we randomly select 30 tasks from three datasets in CUB, Cars, and Places, measuring the similarity between every two tasks after a data augmentation for each sample. The final results are shown in Table \ref{distance_table} and Fig. \ref{distance_figure}.

Table \ref{distance_table} shows the average, maximum and minimum distance in each group of tasks. Each number in the table is computed 10 times, and the mean results are reported. The tasks from the same domain don't have essential differences with the tasks from the other domains. 
The Max and Min values show that the similarity of tasks is not totally related with domain. The tasks which are most similar may not be consist in the same domain.
The Average distance and the Max distance also show that the cross-domain tasks are more likely to be unsimilar to each other. For example, the tasks in CUB domain have a average similarity of 62.05, while similarity of tasks in CUB and other domains are larger. These reasons provide the probability to select training tasks from the other domains.

Fig. \ref{distance_figure} also explains similar property as Table \ref{distance_table}. Tasks from the same domain are sometimes less similar than cross-domain tasks. Even worse, we also observe some tasks, which show higher self divergences than either intra-domain or cross-domain tasks. These tasks have lower accuracy for a classifier that trained with the random training set because the feature or distribution of these tasks is different from the training set. This means that if we randomly select training tasks, the distance between the training tasks and the target tasks is expected to be larger than the minimum distance. As we introduced above, some training tasks are unsimilar to each other, the training process may not converge to a fixed optimal result. This may cause the parameters of the model to fluctuate largely. Therefore, random tasks selection may cause high uncertainty during training. By incorporating the OTTS method, the distance between training and testing tasks can be minimized. Therefore, the optimized training tasks are more likely to be similar with the testing tasks, and guarantee a stable training process. The detail is shown in the next section.

\begin{figure}[!t]
\centering
\includegraphics[width=250pt]{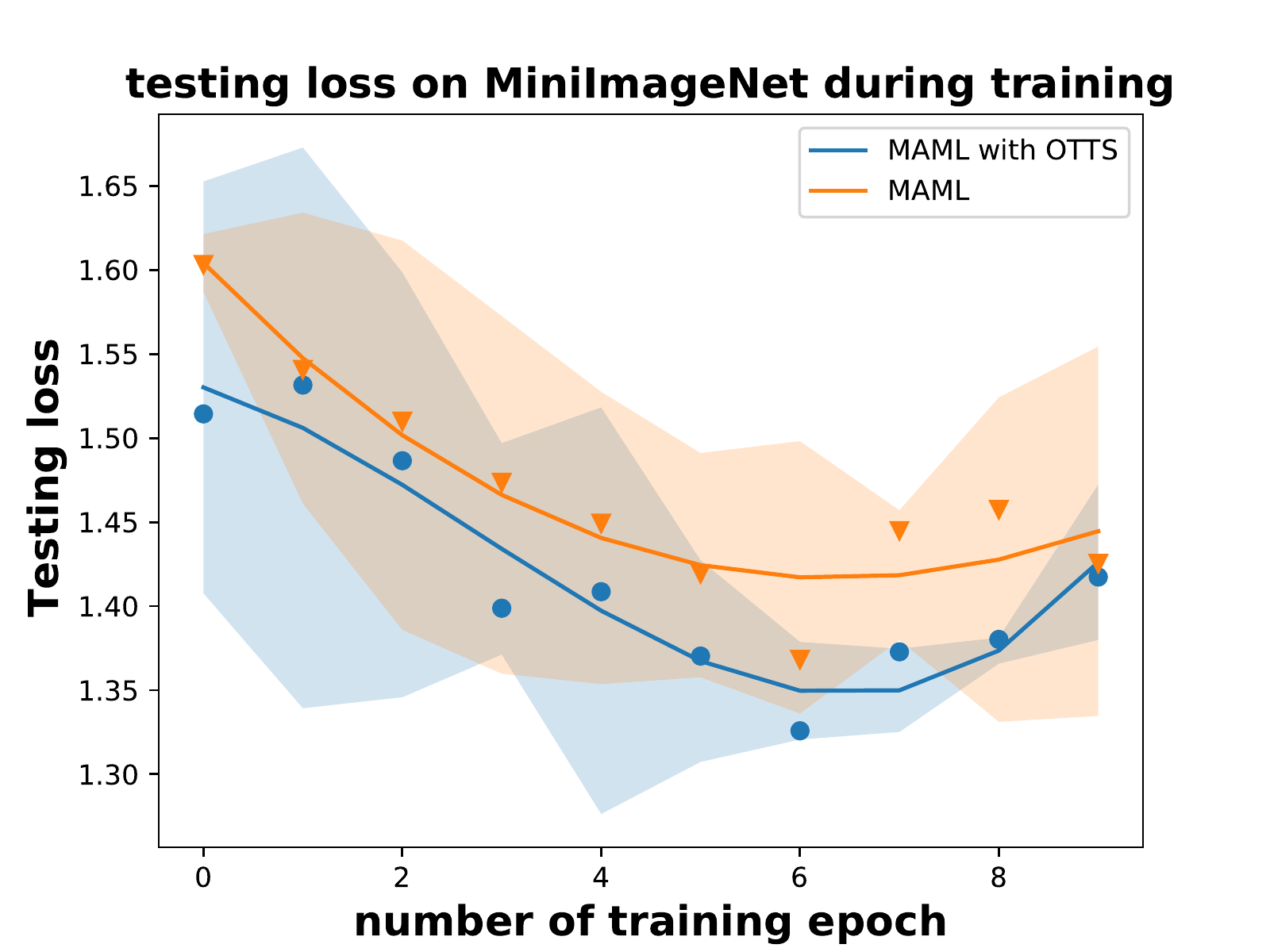}
\caption{\textbf{Testing loss}: This figure shows the change curve of the testing loss during the training process of MAML. The blue curve is the loss from the model which trained with tasks selected by OTTS while the orange one is random tasks. The shadow areas represent the standard deviation of the loss in each epoch. The dots and triangles show the average loss in each epoch.}
\label{loss_curve_figure}
\end{figure}

\begin{table*}[!t]
\renewcommand{\arraystretch}{1.8}
\caption{Testing Accuracy and Loss during training.}
\label{loss_table}
\centering
\begin{tabular}{cccccccccc}
\toprule
MiniImageNet&\multicolumn{2}{c}{Random tasks}&\multicolumn{2}{c}{OTTS tasks}    &CIFAR      &\multicolumn{2}{c}{Random tasks}&\multicolumn{2}{c}{OTTS tasks}\\
\cmidrule(r){2-3} \cmidrule(l){4-5} \cmidrule(l){7-8} \cmidrule(l){9-10}
set ID          &Accuracy(\%)    &Loss        &Accuracy(\%)     &Loss   & set ID    &Accuracy(\%)    &Loss        &Accuracy(\%)     &Loss\\ 
\midrule
set 1              &39.62          &1.2561         &\textbf{39.80}          &\textbf{1.2437}    &set 1       &38.43      &\textbf{1.3729}         &\textbf{44.92}      &1.3732\\
set 2              &38.16          &1.4440         &\textbf{38.70}          &\textbf{1.4323}    &set 2       &37.77      &1.2415         &\textbf{39.40}      &\textbf{1.1581}\\
set 3              &35.06          &1.5868         &\textbf{36.90}          &\textbf{1.5045}    &set 3       &\textbf{39.62}      &\textbf{1.2987}         &39.43      &1.3082\\
set 4              &\textbf{39.28}          &1.3402         &37.06          &\textbf{1.3116}    &set 4       &38.48      &\textbf{1.4923}         &\textbf{40.40}      &1.6477\\
set 5              &\textbf{36.99}          &1.5147         &36.35          &\textbf{1.3561}    &set 5       &39.04      &\textbf{1.4449}         &\textbf{41.26}      &1.4872\\
set 6              &\textbf{37.50}          &1.6423         &37.26          &\textbf{1.5038}    &set 6       &\textbf{36.13}      &\textbf{1.4819}         &35.03      &1.5656\\
set 7              &37.06          &\textbf{1.4942}         &\textbf{39.23}          &1.5167    &set 7       &\textbf{40.06}      &1.4282         &39.45      &\textbf{1.3294}\\
set 8              &35.77          &1.4184         &\textbf{36.06}          &\textbf{1.4020}    &set 8       &32.93      &1.5638         &\textbf{37.00}      &\textbf{1.4685}\\
set 9              &38.30          &1.4574         &\textbf{39.14}          &\textbf{1.3955}    &set 9       &34.10      &1.4169         &\textbf{34.86}      &\textbf{1.2712}\\
set 10             &34.96          &1.5940         &\textbf{36.52}          &\textbf{1.4436}    &set 10      &37.06      &1.5426         &\textbf{39.40}      &\textbf{1.3850}\\
\midrule
total               &37.27          &1.4784         &\textbf{37.70}         &\textbf{1.4110}    &total       &37.36      &1.4284         &\textbf{39.12}      &\textbf{1.3994}\\
\bottomrule
\end{tabular}
\end{table*}

\subsubsection{How similar tasks influence the training process}

In this experiment, we compare how similar tasks selected by OTTS influence the training process of MAML. Particularly, we compare the training processes by employing the randomly selected tasks and the OTTS selected tasks. The results of this experiment are shown in Table \ref{loss_table} and Fig. \ref{loss_curve_figure}.

Table \ref{loss_table} represents the accuracy and the loss with different testing sets from MiniImageNet of CIFAR. Each testing set includes 5 testing tasks that are similar to each other. The training set includes 400 random tasks or selected tasks which are similar to the testing sets. Besides, Table 2 shows that the OTTS can benefit the training process of MAML. In the Table \ref{loss_table}, we can find out that the selected tasks can achieve lower loss value in most cases. For the total testing set which contains all 10 sets, our OTTS outperforms the random selection strategy in both metrics, i.e., accuracy and the loss. Because the testing sets from CIFAR have larger differences with training tasks, the improvements of both accuracy and loss value are also larger. This proved that OTTS can benefit training process even though the tasks are from different domains, and the model is not specially designed for cross domain problems. The selected tasks can achieve higher results especially since the domains are different are whole but contains similar tasks.

Fig. \ref{loss_curve_figure} reflects the change of loss during the training process of MAML. Owing to the OTTS, the loss attenuation is faster than using the random tasks. In the meantime, selected tasks can make the training process more stable which can be seen as the shadow area of the Fig. \ref{loss_curve_figure}. It can be noticed that the blue shadow area shrinks fast during the training process while the orange shadow doesn't follow similar law. This represents that the selected tasks can cause a smoother learning process than random tasks.

These results validate the effectiveness of OTTS with MAML. It can be seen that OTTS can improve the stability of the training process. The multiple examinations validate the stability of OTTS. Selecting multiple training tasks can further reduce the randomness of the task selected by OTTS. During the training process, the training loss converged more stable, and the final loss is lower in most cases by using OTTS selected tasks. In the next section, we will discuss the effect of OTTS on the metric-based model on Cross-domain problems \cite{Cross_domain,Cross_domain_2}.

\begin{figure}[!t]
\centering
\includegraphics[width=250pt]{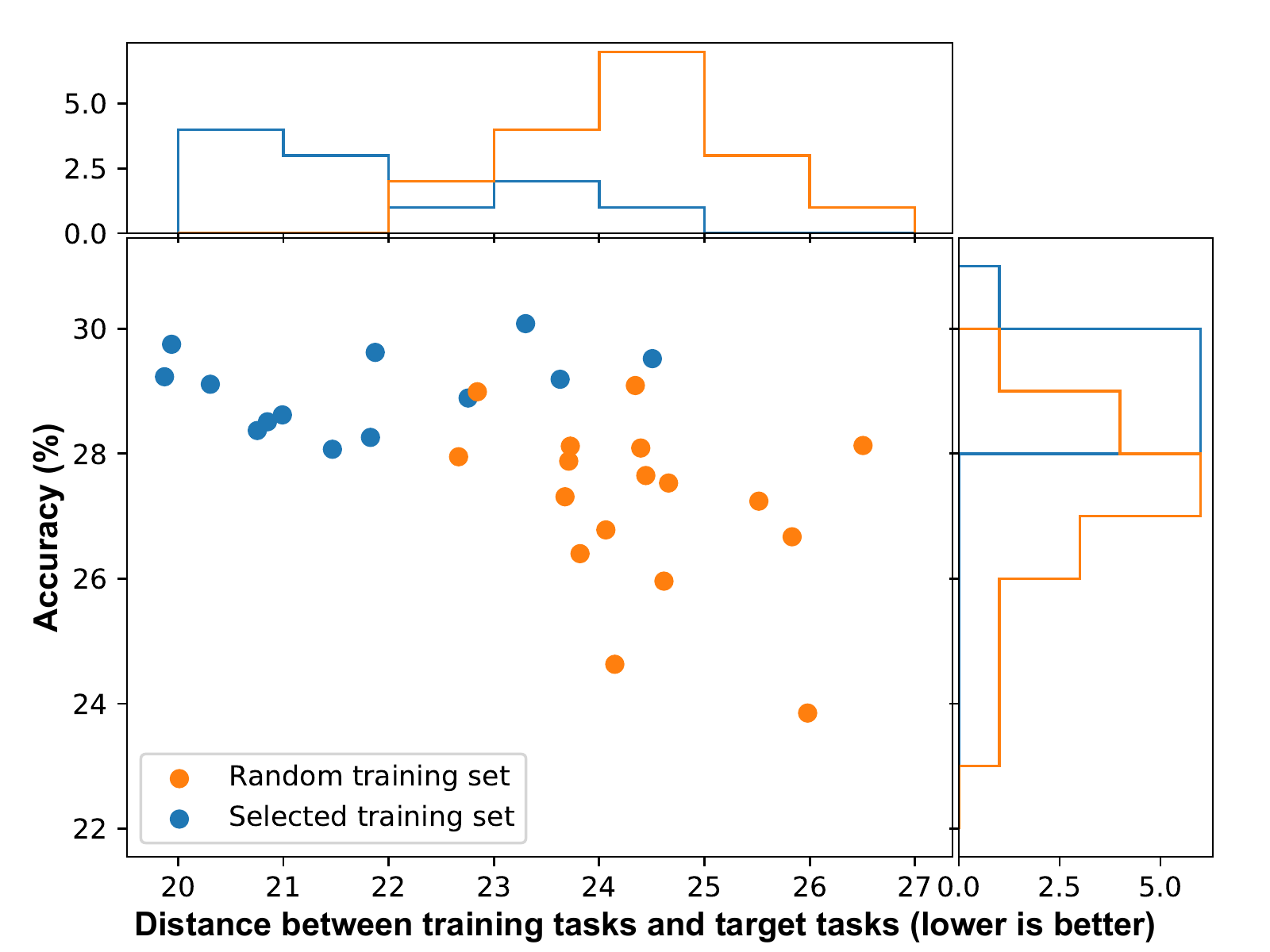}
\caption{\textbf{The relationship between the distance and accuracy}: This figure shows the relationship between the distance of target testing tasks and the source support tasks. The dots in the figure represent the accuracy and distance of each time of the training. The curves denote the frequency of distance and accuracy in each interval.}
\label{distribution_accuracy}
\end{figure}

\begin{table*}[!t]
\renewcommand{\arraystretch}{1.8}
\caption{The accuracy comparison between random strategy and our OTTS algorithm on metric-based models.}
\label{acc_table}
\centering
\begin{tabular}{ccccccc}
\toprule
\multirow{2}{*}{model}&\multicolumn{2}{c}{Cars}&\multicolumn{2}{c}{CUB}&\multicolumn{2}{c}{Places}\\
\cmidrule(r){2-3} \cmidrule(r){4-5} \cmidrule(r){6-7}
                    &random    &OTTS        &random     &OTTS    &random     &OTTS \\ 
\midrule
matchingnet        &24.74\% +- 0.33\%      &27.70\% +- 0.42\%   &26.91\% +- 0.39\%  &31.31\% +- 0.41\% &30.74\% +- 0.47\%        &30.11\% +- 0.45\%\\
protonet          &25.04\% +- 0.37\%    &26.00\% +- 0.37\%      &27.36\% +- 0.37\%
    &28.23\% +- 0.36\%      &27.79\% +- 0.40\%   &29.56\% +- 0.47\%\\
\bottomrule
\end{tabular}
\end{table*}

\begin{table}[!t]
\renewcommand{\arraystretch}{1.8}
\caption{The accuracy of different similarity settings on Cars.}
\label{melting}
\centering
\begin{tabular}{cccc}

\toprule
        &Ss-Ts      &Ss-Tt      &Accuracy\\
\midrule
matchingnet&\checkmark&&25.22\% +- 0.35\%\\
-&\checkmark&\checkmark&25.77\% +- 0.33\%\\
-&&\checkmark&27.70\% +- 0.42\%\\
protonet&\checkmark&&24.71\% +- 0.33\%\\
-&\checkmark&\checkmark&25.48\% +- 0.36\%\\
-&&\checkmark&26.00\% +- 0.37\%\\
\bottomrule
\end{tabular}
\end{table}

\subsubsection{Cross-Domain Few-Shot experiment}

The Cross-Domain Few-Shot experiment selects the tasks from MiniImageNet as training set, while the target tasks are from CUB, Cars, or Places. In this experiment, similar tasks or random tasks are utilized to show the effect of OTTS in Cross-Domain learning. The models used in this experiment are Protonet and Matchingnet. The training strategy is FT. The final results are reported in Table \ref{acc_table}, Fig. \ref{distribution_accuracy} and Table \ref{melting}.

In Table \ref{acc_table}, the accuracy of each model with OTTS selected training set on Cars, Cub, and Places are higher than the model with a random training set. This elaborates that OTTS can select a training set from the source domain to assist the training for the target domain. By using OTTS, the accuracy of the models increase by 1.72\% in average.

Fig. \ref{distribution_accuracy} shows the distance relationship between the source support (Ss) tasks and target testing (Tt) tasks. The "\checkmark" in the table represents the distance relationship is limited otherwise with no limits. The source support tasks are selected according to the target support (Ts) tasks. By selecting the source support tasks, the distance between the target testing tasks and the source support tasks can be partly minimized.  Additionally, the OTTS selected training set preserves a smaller distribution distance than the random selection strategy. Therefore, the accuracy is higher with the tasks selected by OTTS. It is worth mentioning that the OTTS selected tasks can also deliver more stable distance and accuracy during training.


Furthermore, we investigate the relationship between task similarity and classification accuracy in Table \ref{melting}. "Ss-Ts" means that we limit the distance between the source support (Ss) tasks and the target support (Ts) tasks, while the "Ss-Tt" is source support (Ss) tasks with target testing (Tt) tasks. It can be recognized that the similarity between the source support (Ss) tasks and target testing (Tt) tasks can largely improve the accuracy of the model. But the similarity between source support (Ss) tasks with target support (Ts) tasks is not significant and would cause a negative impact when we limit both similarities.


In the cross-domain experiments, we prove the effectiveness of OTTS in cross-domain learning. By using OTTS, the tasks from the source domain can be selected to construct a smaller training set which is more efficient for training. The large average distance between sets can be reduced the accuracy of target testing sets. The usage in cross-domain learning area is also a strength of OTTS than the transfer teacher methods, because OTTS can directly optimize the distribution of the training samples.

\section{Conclusion}
Few-Shot learning problems have received wide attention in recent years aiming to solve a problem that not enough samples can be used in the learning process. Therefore, the Few-Shot problem require extensive samples from other tasks to train the model. However, the samples from other tasks usually have different distribution with the task we wanted to train. Although there are multiple methods to minimize the differences of the tasks, like data augmentation and domain alignment, these methods require either the distribution or the amount of the of samples.
To solve these problems, we proposed a novel method, i.e., OTTS, to minimize the differences by selecting similar tasks from the known datasets. In this method, we employed the self-supervised training method and graph optimal transport distance to measure the similarity between tasks. By using OTTS in multiple experiments, we concluded the properties of this method and proposed the application of the OTTS in Cross-Domain Few-Shot classification problems. These experiments validated the effectiveness of OTTS and achieved better results than the baseline methods. Furthermore, OTTS algorithm can not only select the tasks which are most similar to the target tasks but also be employed in a training strategy that use the tasks with different similarities efficiently. We are also looking forward to the further usages of the task selecting methods, e.g., the training process of OTTS can be carried out simultaneously with the training process of classifier.


%

\ifCLASSOPTIONcaptionsoff
  \newpage
\fi



%

\bibliographystyle{IEEEtran}
\bibliography{sample}

\begin{thebibliography}{10}
\providecommand{\url}[1]{#1}
\csname url@samestyle\endcsname
\providecommand{\newblock}{\relax}
\providecommand{\bibinfo}[2]{#2}
\providecommand{\BIBentrySTDinterwordspacing}{\spaceskip=0pt\relax}
\providecommand{\BIBentryALTinterwordstretchfactor}{4}
\providecommand{\BIBentryALTinterwordspacing}{\spaceskip=\fontdimen2\font plus
\BIBentryALTinterwordstretchfactor\fontdimen3\font minus
  \fontdimen4\font\relax}
\providecommand{\BIBforeignlanguage}[2]{{%
\expandafter\ifx\csname l@#1\endcsname\relax
\typeout{** WARNING: IEEEtran.bst: No hyphenation pattern has been}%
\typeout{** loaded for the language `#1'. Using the pattern for}%
\typeout{** the default language instead.}%
\else
\language=\csname l@#1\endcsname
\fi
#2}}
\providecommand{\BIBdecl}{\relax}
\BIBdecl

\bibitem{few_shot_survey}
Y.~Wang, Q.~Yao, J.~T. Kwok, and L.~M. Ni, ``Generalizing from a few examples:
  A survey on few-shot learning,'' \emph{ACM Computing Surveys (CSUR)},
  vol.~53, no.~3, pp. 1--34, 2020.

\bibitem{protonet}
J.~Snell, K.~Swersky, and R.~S. Zemel, ``Prototypical networks for few-shot
  learning,'' \emph{arXiv preprint arXiv:1703.05175}, 2017.

\bibitem{TKDE_few_shot}
J.~Li, B.~Chiu, S.~Feng, and H.~Wang, ``Few-shot named entity recognition via
  meta-learning,'' \emph{IEEE Transactions on Knowledge and Data Engineering},
  pp. 1--1, 2020.

\bibitem{Data_Augmentation}
Z.~Zhong, L.~Zheng, G.~Kang, S.~Li, and Y.~Yang, ``Random erasing data
  augmentation,'' in \emph{Proceedings of the AAAI Conference on Artificial
  Intelligence}, vol.~34, no.~07, 2020, pp. 13\,001--13\,008.

\bibitem{Data_Augmentation_survey}
C.~Shorten and T.~M. Khoshgoftaar, ``A survey on image data augmentation for
  deep learning,'' \emph{Journal of Big Data}, vol.~6, no.~1, pp. 1--48, 2019.

\bibitem{domain_generalization_survey}
K.~Zhou, Z.~Liu, Y.~Qiao, T.~Xiang, and C.~C. Loy, ``Domain generalization: A
  survey,'' \emph{arXiv preprint arXiv:2103.02503}, 2021.

\bibitem{GOT_DA}
L.~Chen, Z.~Gan, Y.~Cheng, L.~Li, L.~Carin, and J.~Liu, ``Graph optimal
  transport for cross-domain alignment,'' in \emph{International Conference on
  Machine Learning}.\hskip 1em plus 0.5em minus 0.4em\relax PMLR, 2020, pp.
  1542--1553.

\bibitem{task2vec}
A.~Achille, M.~Lam, R.~Tewari, A.~Ravichandran, S.~Maji, C.~C. Fowlkes,
  S.~Soatto, and P.~Perona, ``Task2vec: Task embedding for meta-learning,'' in
  \emph{Proceedings of the IEEE/CVF International Conference on Computer
  Vision}, 2019, pp. 6430--6439.

\bibitem{task2vec_2}
B.~Wallace, Z.~Wu, and B.~Hariharan, ``Can we characterize tasks without labels
  or features?'' in \emph{Proceedings of the IEEE/CVF Conference on Computer
  Vision and Pattern Recognition}, 2021, pp. 1245--1254.

\bibitem{task_distribution_OT}
D.~Alvarez-Melis and N.~Fusi, ``Geometric dataset distances via optimal
  transport,'' \emph{arXiv preprint arXiv:2002.02923}, 2020.

\bibitem{ot}
G.~Peyr{\'e}, M.~Cuturi \emph{et~al.}, ``Computational optimal transport: With
  applications to data science,'' \emph{Foundations and Trends{\textregistered}
  in Machine Learning}, vol.~11, no. 5-6, pp. 355--607, 2019.

\bibitem{ot_graph}
V.~Titouan, N.~Courty, R.~Tavenard, and R.~Flamary, ``Optimal transport for
  structured data with application on graphs,'' in \emph{International
  Conference on Machine Learning}.\hskip 1em plus 0.5em minus 0.4em\relax PMLR,
  2019, pp. 6275--6284.

\bibitem{CL_1}
W.~Huang, J.~Liu, T.~Li, T.~Huang, S.~Ji, and J.~Wan, ``Feddsr: Daily schedule
  recommendation in a federated deep reinforcement learning framework,''
  \emph{IEEE Transactions on Knowledge and Data Engineering}, pp. 1--1, 2021.

\bibitem{CL_2}
W.~Wang, G.~Xu, W.~Ding, Y.~Huang, G.~Li, J.~Tang, and Z.~Liu, ``Representation
  learning from limited educational data with crowdsourced labels,'' \emph{IEEE
  Transactions on Knowledge and Data Engineering}, pp. 1--1, 2020.

\bibitem{CL_3}
B.~Han, I.~W. Tsang, X.~Xiao, L.~Chen, S.-F. Fung, and C.~P. Yu,
  ``Privacy-preserving stochastic gradual learning,'' \emph{IEEE Transactions
  on Knowledge and Data Engineering}, vol.~33, no.~8, pp. 3129--3140, 2021.

\bibitem{CL_survey}
X.~Wang, Y.~Chen, and W.~Zhu, ``A survey on curriculum learning,'' \emph{IEEE
  Transactions on Pattern Analysis and Machine Intelligence}, pp. 1--1, 2021.

\bibitem{self_paced}
X.~Guo, X.~Liu, E.~Zhu, X.~Zhu, M.~Li, X.~Xu, and J.~Yin, ``Adaptive self-paced
  deep clustering with data augmentation,'' \emph{IEEE Transactions on
  Knowledge and Data Engineering}, vol.~32, no.~9, pp. 1680--1693, 2019.

\bibitem{self_paced_2}
Y.~Tang, Y.~Xie, X.~Yang, J.~Niu, and W.~Zhang, ``Tensor multi-elastic kernel
  self-paced learning for time series clustering,'' \emph{IEEE Transactions on
  Knowledge and Data Engineering}, 2019.

\bibitem{transfer_teacher_1}
\BIBentryALTinterwordspacing
D.~Weinshall, G.~Cohen, and D.~Amir, ``Curriculum learning by transfer
  learning: Theory and experiments with deep networks,'' in \emph{Proceedings
  of the 35th International Conference on Machine Learning}, ser. Proceedings
  of Machine Learning Research, J.~Dy and A.~Krause, Eds., vol.~80.\hskip 1em
  plus 0.5em minus 0.4em\relax PMLR, 10--15 Jul 2018, pp. 5238--5246. [Online].
  Available: \url{https://proceedings.mlr.press/v80/weinshall18a.html}
\BIBentrySTDinterwordspacing

\bibitem{transfer_teacher_2}
\BIBentryALTinterwordspacing
G.~Hacohen and D.~Weinshall, ``On the power of curriculum learning in training
  deep networks,'' in \emph{Proceedings of the 36th International Conference on
  Machine Learning}, ser. Proceedings of Machine Learning Research,
  K.~Chaudhuri and R.~Salakhutdinov, Eds., vol.~97.\hskip 1em plus 0.5em minus
  0.4em\relax PMLR, 09--15 Jun 2019, pp. 2535--2544. [Online]. Available:
  \url{https://proceedings.mlr.press/v97/hacohen19a.html}
\BIBentrySTDinterwordspacing

\bibitem{Reinforce_teacher_1}
A.~Graves, M.~G. Bellemare, J.~Menick, R.~Munos, and K.~Kavukcuoglu,
  ``Automated curriculum learning for neural networks,'' in \emph{international
  conference on machine learning}.\hskip 1em plus 0.5em minus 0.4em\relax PMLR,
  2017, pp. 1311--1320.

\bibitem{Reinforce_teacher_2}
T.~Matiisen, A.~Oliver, T.~Cohen, and J.~Schulman, ``Teacher--student
  curriculum learning,'' \emph{IEEE transactions on neural networks and
  learning systems}, vol.~31, no.~9, pp. 3732--3740, 2019.

\bibitem{WD}
Y.~Rubner, C.~Tomasi, and L.~J. Guibas, ``A metric for distributions with
  applications to image databases,'' in \emph{Sixth International Conference on
  Computer Vision (IEEE Cat. No. 98CH36271)}.\hskip 1em plus 0.5em minus
  0.4em\relax IEEE, 1998, pp. 59--66.

\bibitem{GWD}
G.~Peyr{\'e}, M.~Cuturi, and J.~Solomon, ``Gromov-wasserstein averaging of
  kernel and distance matrices,'' in \emph{International Conference on Machine
  Learning}.\hskip 1em plus 0.5em minus 0.4em\relax PMLR, 2016, pp. 2664--2672.

\bibitem{ot_distribution_1}
B.~Muzellec and M.~Cuturi, ``Generalizing point embeddings using the
  wasserstein space of elliptical distributions,'' in \emph{Proceedings of the
  32nd International Conference on Neural Information Processing Systems},
  2018, pp. 10\,258--10\,269.

\bibitem{ot_distribution_2}
C.~Frogner, F.~Mirzazadeh, and J.~Solomon, ``Learning embeddings into entropic
  wasserstein spaces,'' in \emph{International Conference on Learning
  Representations}, 2018.

\bibitem{SS_survey}
L.~Jing and Y.~Tian, ``Self-supervised visual feature learning with deep neural
  networks: A survey,'' \emph{IEEE transactions on pattern analysis and machine
  intelligence}, 2020.

\bibitem{moco}
K.~He, H.~Fan, Y.~Wu, S.~Xie, and R.~Girshick, ``Momentum contrast for
  unsupervised visual representation learning,'' in \emph{Proceedings of the
  IEEE/CVF Conference on Computer Vision and Pattern Recognition}, 2020, pp.
  9729--9738.

\bibitem{BYOL}
J.-B. Grill, F.~Strub, F.~Altch{\'e}, C.~Tallec, P.~H. Richemond,
  E.~Buchatskaya, C.~Doersch, B.~A. Pires, Z.~D. Guo, M.~G. Azar \emph{et~al.},
  ``Bootstrap your own latent: A new approach to self-supervised learning,''
  \emph{arXiv preprint arXiv:2006.07733}, 2020.

\bibitem{CDFS_1}
Y.~Guo, N.~C. Codella, L.~Karlinsky, J.~V. Codella, J.~R. Smith, K.~Saenko,
  T.~Rosing, and R.~Feris, ``A broader study of cross-domain few-shot
  learning,'' in \emph{European Conference on Computer Vision}.\hskip 1em plus
  0.5em minus 0.4em\relax Springer, 2020, pp. 124--141.

\bibitem{CDFS_2}
T.~Adler, J.~Brandstetter, M.~Widrich, A.~Mayr, D.~Kreil, M.~Kopp,
  G.~Klambauer, and S.~Hochreiter, ``Cross-domain few-shot learning by
  representation fusion,'' \emph{arXiv e-prints}, pp. arXiv--2010, 2020.

\bibitem{transfer}
S.~J. Pan and Q.~Yang, ``A survey on transfer learning,'' \emph{IEEE
  Transactions on Knowledge and Data Engineering}, vol.~22, no.~10, pp.
  1345--1359, 2010.

\bibitem{GPML}
C.~E. Rasmussen, ``Gaussian processes in machine learning,'' in \emph{Summer
  school on machine learning}.\hskip 1em plus 0.5em minus 0.4em\relax Springer,
  2003, pp. 63--71.

\bibitem{Cross_domain}
Q.~Zhang, W.~Liao, G.~Zhang, B.~Yuan, and J.~Lu, ``A deep dual adversarial
  network for cross-domain recommendation,'' \emph{IEEE Transactions on
  Knowledge and Data Engineering}, pp. 1--1, 2021.

\bibitem{Cross_domain_2}
M.~Jiang, P.~Cui, X.~Chen, F.~Wang, W.~Zhu, and S.~Yang, ``Social
  recommendation with cross-domain transferable knowledge,'' \emph{IEEE
  Transactions on Knowledge and Data Engineering}, vol.~27, no.~11, pp.
  3084--3097, 2015.

\bibitem{meta_learning_MiniImageNet}
S.~Ravi and H.~Larochelle, ``Optimization as a model for few-shot learning,''
  \emph{Proceedings ofthe 5th International Conference on Learning
  Representations (ICLR)}, 2016.

\bibitem{CIFAR}
A.~Krizhevsky and G.~Hinton, ``Convolutional deep belief networks on
  cifar-10,'' \emph{Unpublished manuscript}, vol.~40, no.~7, pp. 1--9, 2010.

\bibitem{CUB}
P.~Welinder, S.~Branson, T.~Mita, C.~Wah, F.~Schroff, S.~Belongie, and
  P.~Perona, ``Caltech-ucsd birds 200,'' 2010.

\bibitem{Cars}
J.~Krause, M.~Stark, J.~Deng, and L.~Fei-Fei, ``3d object representations for
  fine-grained categorization,'' in \emph{Proceedings of the IEEE international
  conference on computer vision workshops}, 2013, pp. 554--561.

\bibitem{places}
B.~Zhou, A.~Lapedriza, A.~Khosla, A.~Oliva, and A.~Torralba, ``Places: A 10
  million image database for scene recognition,'' \emph{IEEE transactions on
  pattern analysis and machine intelligence}, vol.~40, no.~6, pp. 1452--1464,
  2017.

\bibitem{matchingnet}
O.~Vinyals, C.~Blundell, T.~Lillicrap, D.~Wierstra \emph{et~al.}, ``Matching
  networks for one shot learning,'' \emph{Advances in neural information
  processing systems}, vol.~29, pp. 3630--3638, 2016.

\bibitem{MAML}
C.~Finn, P.~Abbeel, and S.~Levine, ``Model-agnostic meta-learning for fast
  adaptation of deep networks,'' in \emph{International Conference on Machine
  Learning}.\hskip 1em plus 0.5em minus 0.4em\relax PMLR, 2017, pp. 1126--1135.

\bibitem{resnet}
K.~He, X.~Zhang, S.~Ren, and J.~Sun, ``Deep residual learning for image
  recognition,'' in \emph{Proceedings of the IEEE conference on computer vision
  and pattern recognition}, 2016, pp. 770--778.

\bibitem{adam}
D.~P. Kingma and J.~Ba, ``Adam: A method for stochastic optimization,''
  \emph{arXiv e-prints}, pp. arXiv--1412, 2014.

\bibitem{ex3}
H.-Y. Tseng, H.-Y. Lee, J.-B. Huang, and M.-H. Yang, ``Cross-domain few-shot
  classification via learned feature-wise transformation,'' in
  \emph{International Conference on Learning Representations}, 2020.

\end{thebibliography}

%

\begin{IEEEbiography}[{\includegraphics[width=1in,height=1.25in,clip,keepaspectratio]{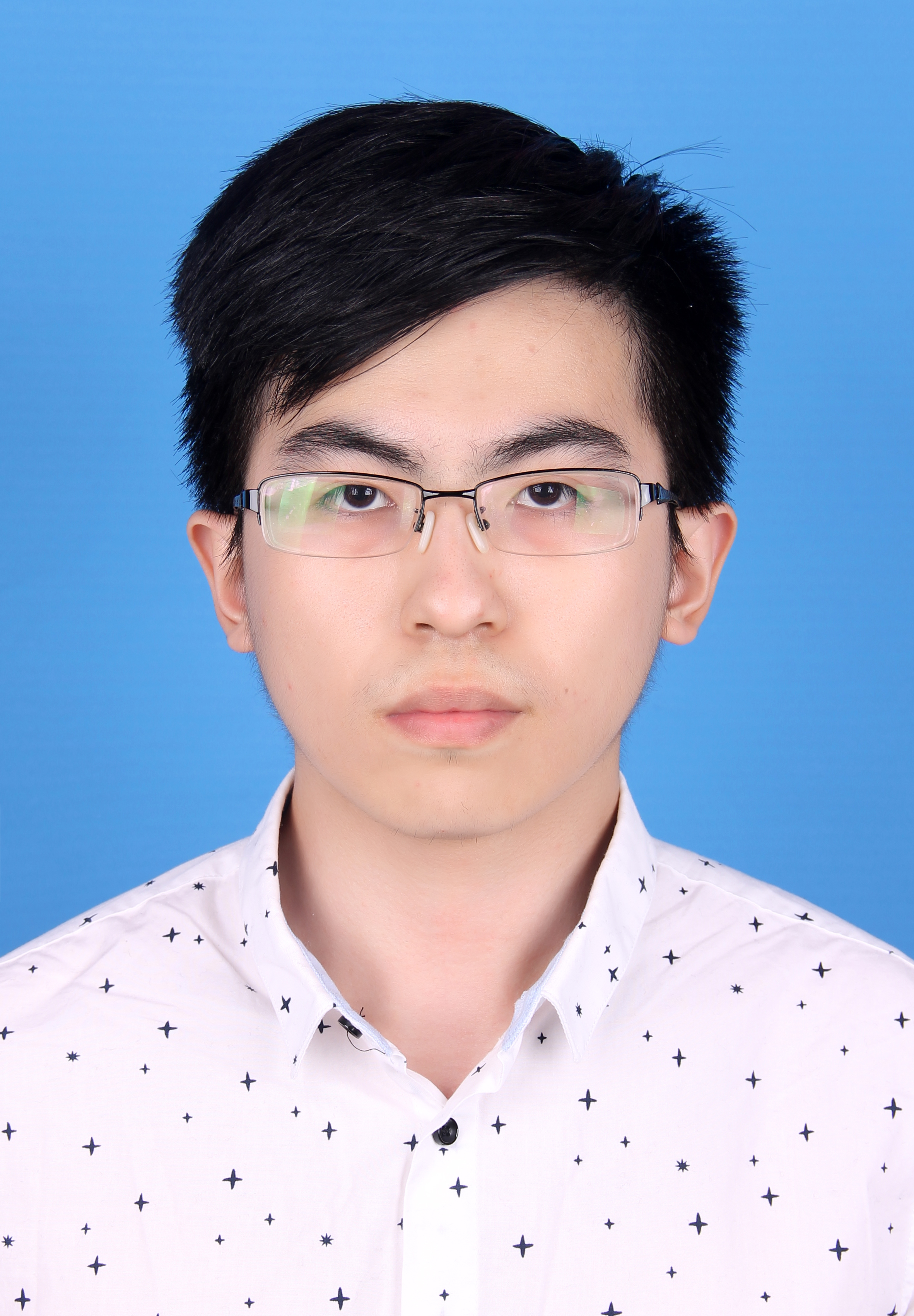}}]{Renjie Xu}
is a master of engineering student in the College of Control Science and Engineering, China University of Petroleum (East China), China. His research interests include meta-learning, Few-Shot learning, and Cross-Domain learning.
\end{IEEEbiography}

\begin{IEEEbiography}[{\includegraphics[width=1in,height=1.25in,clip,keepaspectratio]{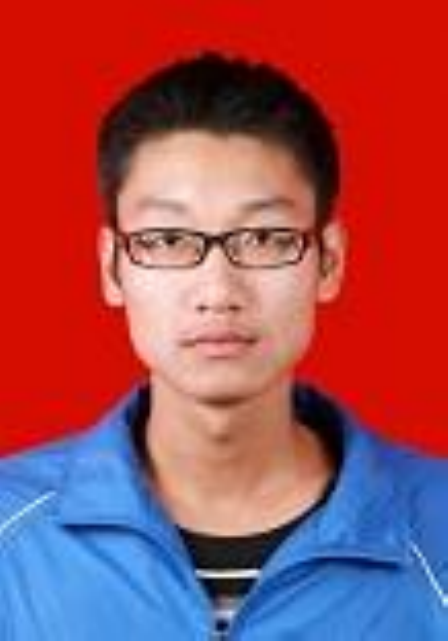}}]{Xinghao Yang}
received the B.Eng. degree in electronic information engineering and M.Eng. degree in information and communication engineering from the China University of Petroleum (East China), Qingdao, China, in 2015 and 2018, respectively. 

His research interests include multi-view learning and adversarial machine learning with publications on IEEE TKDE, AAAI, IJCAI, Information Fusion and Information sciences.
\end{IEEEbiography}

\begin{IEEEbiography}[{\includegraphics[width=1in,height=1.25in,clip,keepaspectratio]{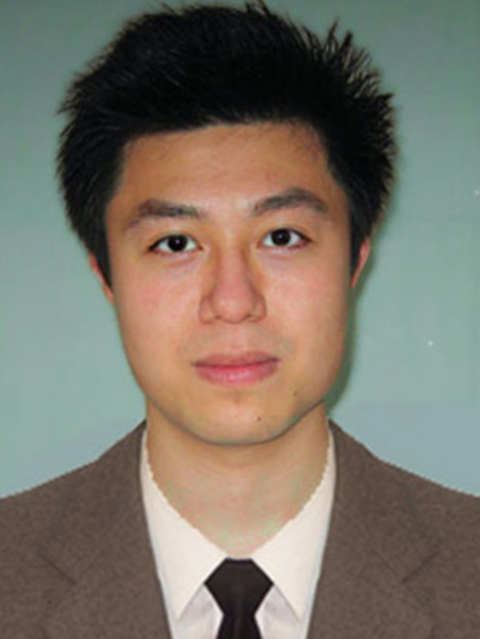}}]{Baodi Liu}
received the Ph.D. degree in Electronic Engineering from Tsinghua University. Currently, he is an assistant professor in College of Information and
Control Engineering, China University of Petroleum, China. His research interests include computer vision and machine learning.
\end{IEEEbiography}

\begin{IEEEbiography}[{\includegraphics[width=1in,height=1.25in,clip,keepaspectratio]{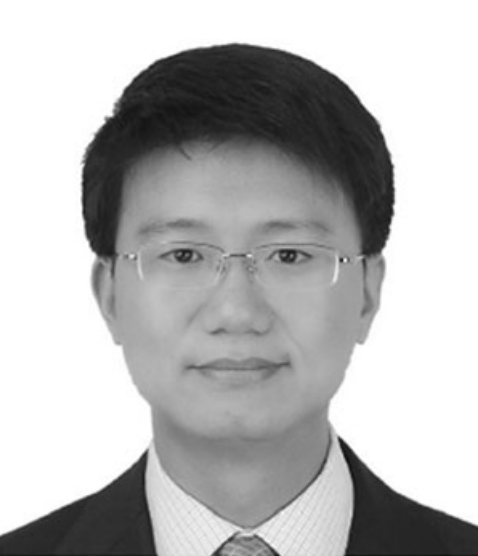}}]{Kai Zhang}
received the Ph.D. degree in petroleum engineering from the China
University of Petroleum (East China), Qingdao, China, in 2008.

From June 2007 to May 2008, he studied with the University of Tulsa, Tulsa, OK, USA. He has been a Teacher with the China University of Petroleum (East China) since 2008. He teaches courses, including fluid flow in porous media and reservoir engineering. As a Project Leader, he has been in charge of three projects supported by the Natural Science Foundation of China, one project supported by the National Natural Science Foundation of Shandong Province, and 20 projects supported by SINOPEC, CNOOC, and CNPC. He has already published more than 60 papers. His research focuses on reservoir simulation, production optimization, history matching, and development of nonconventional reservoir.
\end{IEEEbiography}

\begin{IEEEbiography}[{\includegraphics[width=1in,height=1.25in,clip,keepaspectratio]{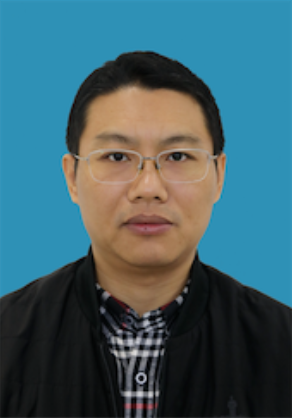}}]{Weifeng Liu}
 (M’12-SM’17) is currently a Professor with the College of Control Science and Engineering, China University of Petroleum (East China), China. He received the double B.S. degree in automation and business administration and the Ph.D. degree in pattern recognition and intelligent systems from the University of Science and Technology of China, Hefei, China, in 2002 and 2007, respectively. His current research interests include pattern recognition and machine learning. He has authored or co-authored a dozen papers in top journals and prestigious conferences including 10 ESI Highly Cited Papers and 3 ESI Hot Papers. Dr. Weifeng Liu serves as associate editor for Neural Processing Letter, co-chair for IEEE SMC technical committee on cognitive computing, and guest editor of special issue for Signal Processing, IET Computer Vision, Neurocomputing, and Remote Sensing. He also serves dozens of journals and conferences.
\end{IEEEbiography}







\end{document}